\documentclass[11pt]{article}

\usepackage[preprint]{acl}

\usepackage{times}
\usepackage{latexsym}

\usepackage[T1]{fontenc}

\usepackage[utf8]{inputenc}

\usepackage{microtype}

\usepackage{inconsolata}

\usepackage{graphicx}

\usepackage{amsmath}
\usepackage{amssymb}

\usepackage{booktabs}
\usepackage{multirow}
\usepackage{makecell}
\usepackage[table]{xcolor}
\usepackage{graphicx}

\usepackage{algorithm}
\usepackage{algpseudocode}

\usepackage{adjustbox}

\newcommand{\pos}[1]{\textcolor{green!30!black}{#1}}
\newcommand{\negv}[1]{\textcolor{red!70!black}{#1}}

\title{NeuPAT: Neuron-aware Plasticity Allocation Tuning for Language-Preserving MLLMs }

\author{
 \textbf{Jiayue Jin\textsuperscript{1,2}},
 \textbf{Jingwei Zhang\textsuperscript{2,5}},
 \textbf{Chen Wang\textsuperscript{2,6}},
 \textbf{Jing Liu\textsuperscript{2,3,4}},
 \textbf{Longteng Guo\textsuperscript{2,3,4*}}\\
\\
 \textsuperscript{1}College of Intelligent Robotics and Advanced Manufacturing, Fudan University\\
 \textsuperscript{2}Zhongguancun Academy,
 \textsuperscript{3}Institute of Automation, Chinese Academy of Sciences\\
 \textsuperscript{4}School of Artificial Intelligence, University of Chinese Academy of Sciences\\
 \textsuperscript{5}Tianjin University,
 \textsuperscript{6}Nankai University
 \\
 \small{
   \textbf{Correspondence:} \href{mailto:longteng.guo@nlpr.ia.ac.cn}{longteng.guo@nlpr.ia.ac.cn}
 }
}

\begin{document}
\maketitle

\begin{abstract}
Multimodal expansion of large language models (LLMs) enables new perceptual capabilities but often compromises the language intelligence acquired during pretraining. In this work, we investigate this phenomenon from the perspective of internal adaptation dynamics and discover that neurons in pretrained LLMs exhibit heterogeneous plasticity during multimodal learning: some neurons are critical for preserving language capabilities, while others are more adaptive to multimodal knowledge. Based on this insight, we propose NeuPAT (Neuron-aware Plasticity Allocation Tuning), a lightweight and architecture-agnostic framework that allocates neuron-wise update constraints during multimodal instruction tuning. NeuPAT uses a small-scale probing stage to estimate neuron adaptation patterns and selectively protects language-sensitive neurons while promoting multimodal adaptation through more plastic neurons. Experiments across diverse LLM families demonstrate that NeuPAT recovers 94.5\% of the language capability degradation caused by vanilla tuning on 11 language benchmarks while maintaining comparable multimodal performance, providing an effective approach for capability-preserving multimodal expansion.
\end{abstract}

\begin{figure}[t]
\centering
\includegraphics[width=0.482\textwidth]{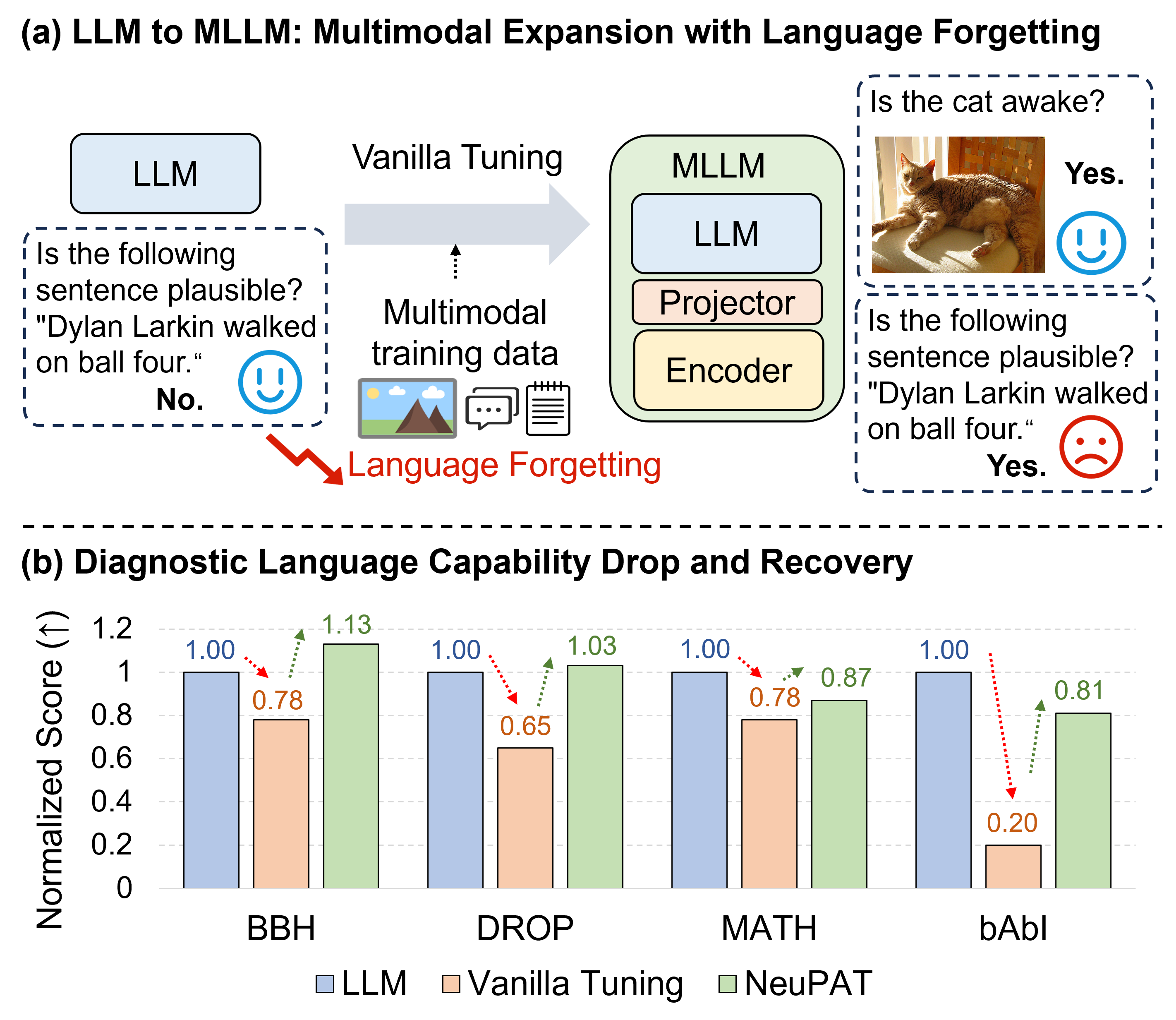}
\caption{\textbf{Pure-text capability degradation and recovery after visual instruction tuning.}
(a) Expanding an LLM into an MLLM with multimodal instruction data can induce language forgetting.
(b) Vanilla tuning degrades performance on representative pure-text benchmarks, whereas NeuPAT recovers most of the lost capability. Scores are normalized by the original LLM performance on each benchmark.}
\label{conceptual}
\end{figure}

\section{Introduction}

Large language models (LLMs) demonstrate strong capabilities in language understanding, reasoning, knowledge acquisition, and code generation through large-scale text pretraining~\cite{brown2020language,chowdhery2023palm,openai2023gpt4,touvron2023llama}.
Extending these models beyond text has driven rapid progress in multimodal large language models (MLLMs)~\cite{liu2023visual,dai2023instructblip,wang2024qwen2,bai2025qwen3,li2024llava,team2025kimi,wang2025internvl3}.
A dominant paradigm preserves a pretrained LLM as the reasoning backbone, connects it to modality-specific encoders via projection modules, and performs multimodal instruction tuning to align new modalities with the language space~\cite{liu2023visual,dai2023instructblip}.
This paradigm enables strong performance on multimodal tasks such as visual question answering and reasoning~\cite{mmbench,mmmu,mmstar}.

Ideally, multimodal expansion should represent capability evolution: the model acquires new multimodal understanding while preserving the language intelligence accumulated during pretraining.
This preservation is fundamental because modern MLLMs remain language-grounded systems, where language capabilities provide the foundation for reasoning, knowledge organization, and multimodal generalization.

However, this ideal preservation is not always achieved.
As shown in Figure~\ref{conceptual}, multimodal instruction tuning often causes substantial degradation on language reasoning benchmarks, with an average performance drop of approximately 39.8\% across key language evaluations.
This reveals that multimodal expansion is not purely additive; adapting the pretrained backbone toward multimodal distributions can interfere with existing language representations and overwrite capabilities essential for language-based reasoning.
Therefore, an important question arises:
\textit{How can we expand the capability boundary of LLMs with new modalities while preserving the language intelligence that forms the foundation of multimodal intelligence?}

Recent studies have explored strategies to alleviate language capability degradation during multimodal expansion.
Existing solutions mainly rely on external interventions, including text-only data replay~\cite{lu2024deepseek, bai2025qwen3}, architectural modifications~\cite{zhang2024wings,wang2025iaa,lu2025genieblue}, and post-training model merging~\cite{ratzlaff2024training,yu2025locate,wang2026plam,li2025dpimerge}.
Text replay retains pretrained abilities through additional language supervision but requires extra data and careful objective balancing.
Architecture-based methods isolate multimodal adaptation from the language backbone at the cost of customized designs and added complexity.
Model merging combines the original LLM and adapted MLLM after training, yet may introduce trade-offs between language preservation and multimodal adaptation.
More importantly, these methods rely on auxiliary interventions rather than directly regulating backbone adaptation during multimodal learning.

In this work, we revisit multimodal expansion from the perspective of internal adaptation dynamics.
We investigate whether the pretrained LLM contains heterogeneous adaptation capacities that can be selectively regulated during multimodal learning.
Our key insight is that neurons within the pretrained backbone exhibit distinct adaptation patterns: some are more critical for preserving language intelligence, while others provide greater flexibility for absorbing new multimodal knowledge.
Therefore, uniformly updating all neurons during multimodal tuning may unnecessarily disrupt language-critical representations while underutilizing the model's intrinsic plasticity.

Based on this insight, we propose \textbf{NeuPAT} (Neuron-aware Plasticity Allocation Tuning), a neuron-level capability-preserving tuning framework for multimodal expansion.
NeuPAT introduces a lightweight probing stage to estimate neuron-wise adaptation characteristics using a small set of diagnostic samples, without introducing additional training data or modifying the model architecture.
Based on these measurements, NeuPAT dynamically allocates update constraints during multimodal instruction tuning: neurons sensitive to language capability preservation are protected from excessive adaptation, while neurons with greater multimodal plasticity are encouraged to acquire new knowledge.
By regulating internal adaptation dynamics, NeuPAT enables capability-preserving expansion from LLMs to MLLMs.

As shown in Figure~\ref{conceptual}, NeuPAT substantially reduces language capability degradation during multimodal expansion, recovering 90.0\% of lost performance across 4 language reasoning benchmarks.
Experiments across 6 LLMs, 11 language benchmarks, and 5 multimodal benchmarks demonstrate that NeuPAT consistently generalizes across model families and scales, providing an efficient and architecture-agnostic solution for preserving language intelligence during multimodal expansion.

Our contributions are summarized as follows:
\begin{itemize}

\item We reveal heterogeneous adaptation dynamics within LLM backbones during multimodal expansion, showing that different neurons exhibit distinct adaptation patterns associated with language preservation and multimodal learning.

\item We propose NeuPAT, a neuron-aware plasticity allocation framework, enabling efficient and architecture-agnostic multimodal expansion while preserving language intelligence.

\item Extensive experiments demonstrate that NeuPAT consistently preserves language intelligence across diverse LLM families and model scales while maintaining comparable multimodal performance, enabling scalable capability expansion from LLMs to MLLMs.

\end{itemize}

\begin{figure*}[t]
\centering
\includegraphics[width=0.99\textwidth]{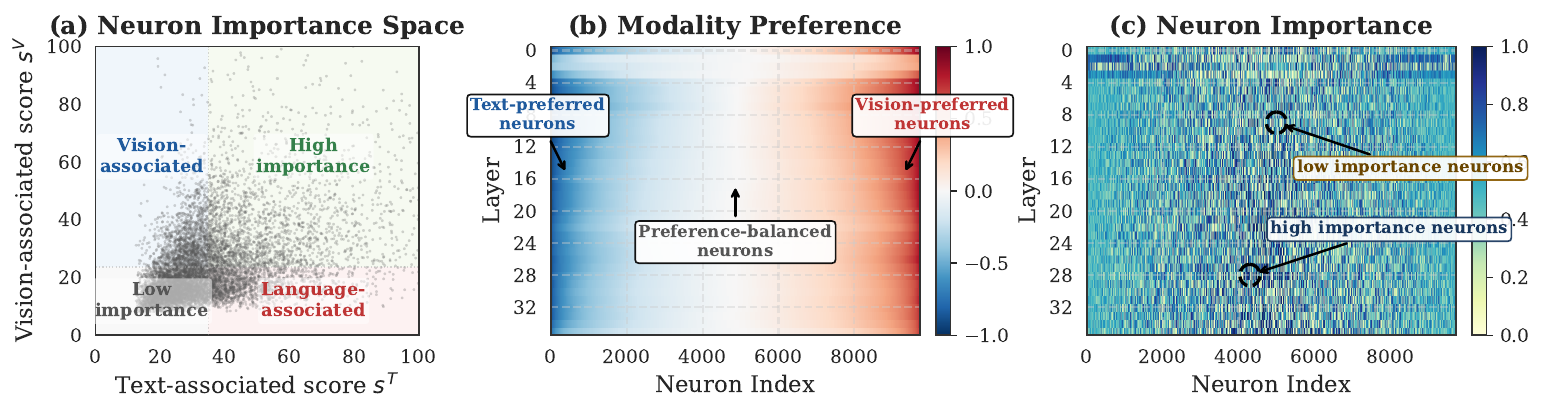} 
\caption{
\textbf{Neuron modality preference and importance during multimodal expansion of LLMs.}
(a) Raw text- and vision-associated importance scores, showing heterogeneous modality associations across neurons.
(b) Layer-wise modality preference $P_{l,u}=v_{l,u}-t_{l,u}$, with neurons sorted by preference within each layer.
(c) Overall importance $I_{l,u}=(v_{l,u}+t_{l,u})/2$ under the same ordering, showing diverse importance levels among neurons with similar preferences.
These results reveal heterogeneous plasticity patterns in the pretrained backbone, motivating neuron-aware plasticity allocation during multimodal adaptation.
}

\label{motivation}
\end{figure*}

\section{Related Work}
\subsection{Multimodal Large Language Models}

MLLMs typically adopt an encoder-connector-LLM architecture, mapping visual features into the language space via cross-attention, Q-Former, or lightweight projection modules~\cite{alayrac2022flamingo,li2023blip,dai2023instructblip,
zhu2024minigpt,liu2023visual,liu2024improved}. Recent models, including the Qwen-VL series, DeepSeek-VL, InternVL3, and LLaVA-OneVision, improve multimodal perception and reasoning with stronger encoders, larger corpora, and advanced post-training~\cite{bai2023qwen,wang2024qwen2,bai2025qwen3,
lu2024deepseek,wang2025internvl3,zhu2025internvl3,li2024llava}. Despite this progress, most MLLMs still adapt pretrained LLM backbones through image-text alignment and visual instruction tuning, potentially degrading their language capabilities and motivating our study.

\subsection{Methods for Mitigating Forgetting}

A common solution is to mix text-only and multimodal data~\cite{lu2024deepseek,bai2023qwen,wang2024qwen2,bai2025qwen3}, which increases training cost and requires careful ratio tuning. Existing alternatives apply general parameter-efficient or continual-learning methods~\cite{hu2022lora,kirkpatrick2017overcoming}, modify the architecture~\cite{zhang2024wings,wang2025iaa,lu2025genieblue}, or merge the adapted model with the original LLM~\cite{ratzlaff2024training,yu2025locate,wang2026plam,li2025dpimerge}. In contrast, NeuPAT leverages neuron modality preferences to preserve language capabilities during multimodal learning without text replay, architectural changes, or post-hoc merging.

\begin{figure*}[t]
\centering
\includegraphics[width=0.99\textwidth]{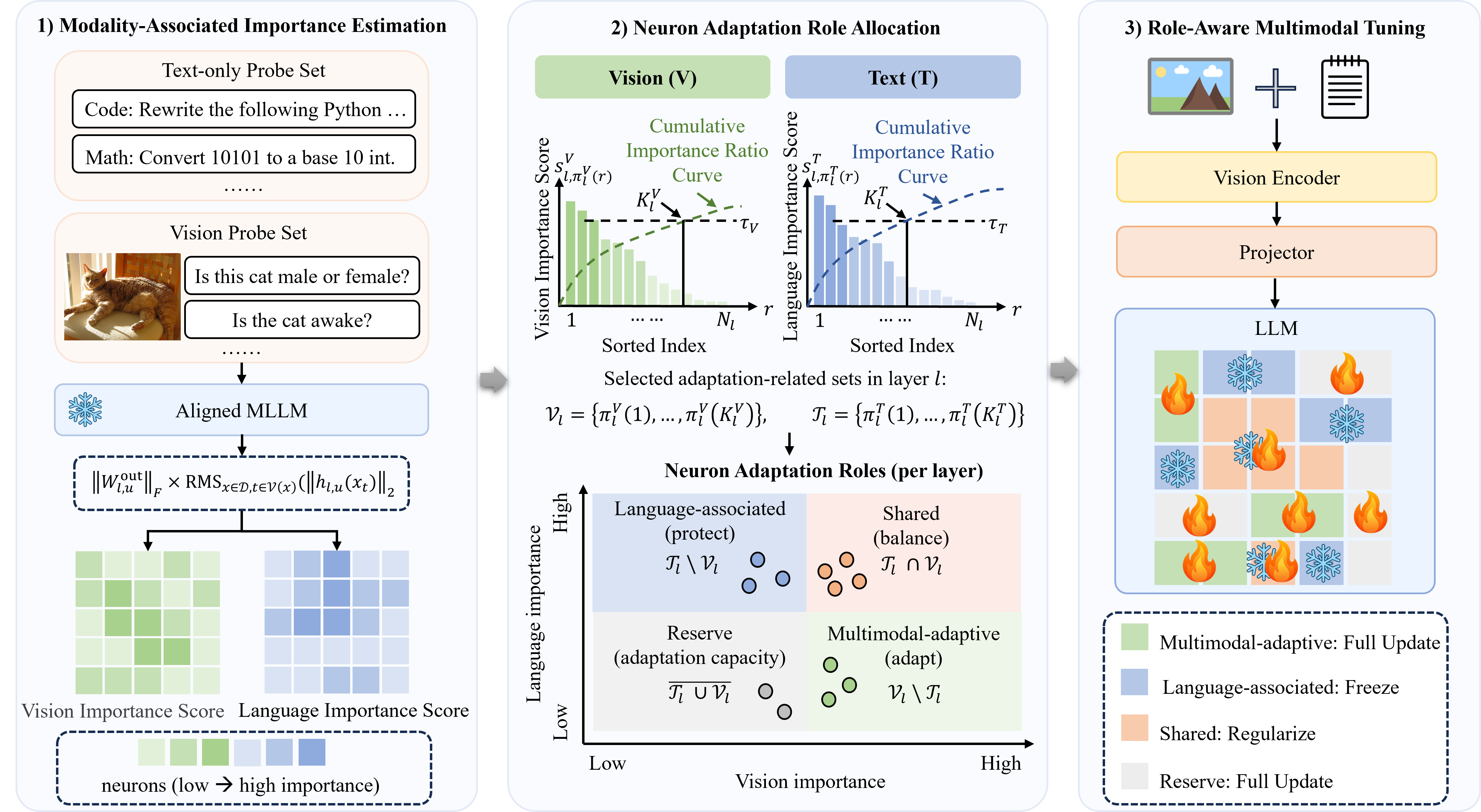} 
\caption{
\textbf{Overview of NeuPAT}.
NeuPAT identifies neuron adaptation roles through lightweight modality probing and assigns role-specific update constraints during multimodal tuning, preserving language intelligence while acquiring multimodal capabilities.
}

\label{main}
\end{figure*}

\section{Neuron-Level Modality Preference Analysis}
\label{sec:observation}
To enable capability-preserving multimodal expansion, we first analyze how different neurons within the pretrained LLM backbone respond to text and visual inputs. 
Our goal is to investigate whether the backbone contains heterogeneous plasticity patterns during multimodal adaptation. 
Without such knowledge, uniform update strategies, such as globally freezing or regularizing the backbone, may preserve language capabilities but unnecessarily restrict the model's ability to acquire new multimodal knowledge.

\subsection{Modality-Associated Neuron Importance Estimation}

For a Transformer layer $l$, we use $u$ to index a neuron, corresponding to an intermediate hidden dimension together with its associated input- and output-side parameter slices. 
Let $h_{l,u}(x_t)$ denote the activation of neuron $u$ at token position $t$, and let $W^{\mathrm{out}}_{l,u}$ denote its output-side parameter slice that writes the neuron response back to the residual stream.

We construct lightweight text-only and visual probing sets, $\mathcal{D}_T$ and $\mathcal{D}_V$, respectively. 
Inspired by activation-aware importance estimation~\cite{sun2024simple}, we define the modality-associated importance score of neuron $u$ under probing set $\mathcal{D}$ as
\begin{equation}
\begin{split}
    s_{l,u}(\mathcal{D})
    ={}&
    \left\|W^{\mathrm{out}}_{l,u}\right\|_F
    \\
    &\cdot
    \mathrm{RMS}_{x\in\mathcal{D},\,t\in\mathcal{V}(x)}
    \left(
    \left\|h_{l,u}(x_t)\right\|_2
    \right),
\end{split}
\label{eq:wanda}
\end{equation}
where $\mathcal{V}(x)$ denotes the valid non-padding token positions. 
This score jointly considers neuron activation strength and its output contribution, providing a proxy for neuron importance under a specific input modality.

We compute the visual- and language-associated importance scores as
\begin{equation}
s^V_{l,u}=s_{l,u}(\mathcal{D}_V),
\qquad
s^T_{l,u}=s_{l,u}(\mathcal{D}_T).
\label{eq:norm}
\end{equation}

For visualization and comparison within each layer, the scores are normalized as
\begin{equation}
    v_{l,u} = \mathrm{Norm}(s^V_{l,u}), \qquad
    t_{l,u} = \mathrm{Norm}(s^T_{l,u}),
\label{eq:ranknorm}
\end{equation}
where $\mathrm{Norm}(\cdot)$ denotes layer-wise normalization.

Based on the normalized importance scores, we define two complementary properties:
\begin{equation}
    P_{l,u} = v_{l,u}-t_{l,u},  \qquad
    I_{l,u} = \frac{v_{l,u}+t_{l,u}}{2},
\end{equation}
where $P_{l,u}$ measures the relative modality preference of a neuron, while $I_{l,u}$ measures its overall importance across both modalities. 
A positive $P_{l,u}$ indicates stronger association with visual inputs, whereas a negative value indicates stronger association with language inputs. 
Values close to zero indicate balanced importance across modalities.

\subsection{Empirical Observations}

Figure~\ref{motivation} illustrates neuron importance patterns from three perspectives. 
Figure~\ref{motivation}(a) compares visual- and language-associated importance scores, revealing heterogeneous modality associations among neurons. 
Figures~\ref{motivation}(b) and~\ref{motivation}(c) further visualize the layer-wise modality preference $P_{l,u}$ and overall importance $I_{l,u}$, respectively, where neurons are sorted according to their preference within each layer.

Based on these analyses, we obtain three observations.

\textbf{1) Heterogeneous modality-associated plasticity.}
Neurons exhibit diverse preferences toward language and visual inputs, indicating that the pretrained backbone does not participate uniformly in multimodal adaptation. 
Consequently, applying identical updates to all neurons may unnecessarily perturb language-critical representations while limiting the utilization of neurons better suited for multimodal adaptation.

\textbf{2) Different neurons require different update constraints.}
Neurons with balanced modality preferences are not homogeneous: some exhibit high importance under both input types, suggesting shared functionality across modalities, while others show relatively low importance and may provide additional flexibility for acquiring new multimodal knowledge. 
This indicates that different neurons require different levels of update constraint rather than uniform protection or adaptation.

\textbf{3) Plasticity patterns vary across layers.}
The distributions of modality preference and importance vary substantially across Transformer layers, indicating that different layers exhibit distinct adaptation patterns. 
Therefore, fixed global allocation strategies may fail to capture layer-specific characteristics.

Together, these observations reveal heterogeneous adaptation requirements within the pretrained backbone: some neurons require protection to preserve language intelligence, while others provide flexibility for multimodal adaptation. 
This motivates neuron-aware allocation of update flexibility during multimodal expansion.

\section{Methods}
NeuPAT performs neuron-aware update allocation during multimodal instruction tuning. 
It estimates modality-associated importance, assigns neurons with different adaptation roles, and applies role-specific update constraints to balance language preservation and multimodal adaptation.
Figure~\ref{main} illustrates the overall pipeline.

\subsection{Modality-Associated Importance Estimation}

Following the analysis in Section~\ref{sec:observation}, NeuPAT first estimates modality-associated importance for each neuron using lightweight probing sets. 
Specifically, we construct visual and text-only probing sets and compute the visual- and language-associated importance scores, $s^V_{l,u}$ and $s^T_{l,u}$, using Eq.~\ref{eq:wanda}. 
These scores measure the contribution of individual neurons under different input modalities and provide the basis for subsequent adaptation role allocation.

For each layer $l$, the importance scores are normalized using layer-wise normalization using Eq.~\ref{eq:norm}.
The normalized scores are used to characterize the modality-associated importance distribution of neurons within each layer.

\subsection{Neuron Adaptation Role Allocation}
NeuPAT identifies neurons with different adaptation roles according to their modality-associated importance distributions. 
For each layer $l$, let 
$s^V_l\in\mathbb{R}^{N_l}$ and 
$s^T_l\in\mathbb{R}^{N_l}$ 
denote the visual- and language-associated importance scores of all $N_l$ neurons.

For each modality $a\in\{V,T\}$, we select the smallest neuron subset whose cumulative importance accounts for a predefined coverage ratio $\tau_a$:
\begin{equation}
    K^a_l
    =
    \min
    \left\{
    K:
    \frac{
    \sum_{r=1}^{K} s^a_{l,\pi^a_l(r)}
    }{
    \sum_{r=1}^{N_l} s^a_{l,\pi^a_l(r)}
    +\epsilon
    }
    \ge
    \tau_a
    \right\},
\end{equation}
where $\pi_l^a(\cdot)$ denotes the neuron indices ranked by modality-associated importance.
The selected neuron sets for visual and language inputs are denoted as
$\mathcal{V}_l$ and $\mathcal{T}_l$, respectively.

Based on the relationship between $\mathcal{V}_l$ and $\mathcal{T}_l$, we derive four neuron adaptation roles:
\begin{equation}
\begin{aligned}
\mathcal{C}^{lang}_l &= \mathcal{T}_l\setminus\mathcal{V}_l,
&
\mathcal{C}^{multi}_l &= \mathcal{V}_l\setminus\mathcal{T}_l,
\\
\mathcal{C}^{shared}_l &= \mathcal{T}_l\cap\mathcal{V}_l,
&
\mathcal{C}^{reserve}_l &= \overline{\mathcal{T}_l\cup\mathcal{V}_l}.
\end{aligned}
\end{equation}
where the complement in $\mathcal{C}^{reserve}_l$ is taken with respect to all neurons in layer $l$. 
Here, $\mathcal{C}^{lang}_l$ contains neurons primarily associated with language inputs and therefore requires protection during multimodal adaptation. 
$\mathcal{C}^{multi}_l$ contains neurons associated with visual inputs and provides adaptive capacity for acquiring multimodal knowledge. 
$\mathcal{C}^{shared}_l$ contains neurons important to both modalities and requires constrained updates, while $\mathcal{C}^{reserve}_l$ provides additional flexibility for multimodal learning.

\begin{table*}[t]
\centering
\caption{
Comparison on 11 language benchmarks grouped into Language and Logical Reasoning Tasks, Math and Code Reasoning Tasks, and General Tasks.
$\Delta(2)\!-\!(1)$ denotes Vanilla Tuning $-$ LLM, while
$\Delta(9)\!-\!(2)$ denotes NeuPAT $-$ Vanilla Tuning.
Bold values indicate the best results among multimodally adapted models.
}
\label{tab:language_results}

\renewcommand{\arraystretch}{1.10}
\setlength{\tabcolsep}{3.0pt}
\fontsize{8pt}{8pt}\selectfont

\begin{adjustbox}{max width=\textwidth}
\begin{tabular}{
cl|
cccc>{\columncolor{gray!5}}c|
ccc>{\columncolor{gray!5}}c|
cccc>{\columncolor{gray!5}}c|
>{\columncolor{gray!8}}c
}
\toprule

\multicolumn{2}{c|}{\multirow{2}{*}{\textbf{Method}}}
& \multicolumn{5}{c|}{\textbf{Language and Logical Reasoning Tasks}}
& \multicolumn{4}{c|}{\textbf{Math and Code Reasoning Tasks}}
& \multicolumn{5}{c|}{\textbf{General Tasks}}
& \multicolumn{1}{c}{\textbf{Overall}} \\

\cmidrule(lr){3-7}
\cmidrule(lr){8-11}
\cmidrule(lr){12-16}
\cmidrule(lr){17-17}

\multicolumn{2}{c|}{}
& \textbf{BBH}
& \textbf{bAbI}
& \textbf{DROP}
& \textbf{LogiQA2}
& \textbf{Avg.}
& \textbf{MATH-500}
& \textbf{MBPP}
& \textbf{GSM8K}
& \textbf{Avg.}
& \textbf{SocialIQA}
& \textbf{CoQA}
& \textbf{GPQA}
& \textbf{ARC-C}
& \textbf{Avg.}
& \textbf{Avg.} \\

\midrule

\textbf{(1)}
& LLM
& 47.83
& 13.20
& 15.30
& 35.94
& 28.07
& 64.60
& 65.80
& 86.20
& 72.20
& 50.10
& 67.53
& 36.38
& 55.97
& 52.50
& 48.99 \\

\textbf{(2)}
& + Vanilla Tuning
& 37.12
& 2.68
& 9.90
& 32.95
& 20.66
& 50.40
& 62.60
& 83.47
& 65.49
& 46.78
& 67.37
& 34.15
& 55.72
& 51.01
& 43.92 \\

\multicolumn{2}{c|}{$\boldsymbol{\Delta(2)\!-\!(1)}$}
& \negv{-10.71}
& \negv{-10.52}
& \negv{-5.40}
& \negv{-2.99}
& \negv{\textbf{-7.41}}
& \negv{-14.20}
& \negv{-3.20}
& \negv{-2.73}
& \negv{\textbf{-6.71}}
& \negv{-3.32}
& \negv{-0.16}
& \negv{-2.23}
& \negv{-0.25}
& \negv{\textbf{-1.49}}
& \negv{\textbf{-5.07}} \\

\midrule

\textbf{(3)}
& + LoRA
& 49.72
& 1.57
& 11.42
& 33.97
& 24.17
& 46.20
& 65.40
& 81.96
& 64.52
& \textbf{51.38}
& \textbf{70.58}
& 36.16
& 55.63
& 53.44
& 45.82 \\

\textbf{(4)}
& + EWC
& 42.50
& 2.74
& 12.90
& 35.88
& 23.51
& 52.80
& 66.00
& 85.52
& 68.11
& 47.65
& 69.77
& 35.71
& \textbf{57.51}
& 52.66
& 46.27 \\

\textbf{(5)}
& + WINGS
& 51.76
& 2.70
& 11.03
& 34.16
& 24.91
& 53.20
& 63.20
& 79.91
& 65.44
& 50.00
& 70.50
& 35.49
& 56.48
& 53.12
& 46.22 \\

\textbf{(6)}
& + TIES
& 38.69
& 2.67
& 14.70
& 35.62
& 22.92
& 51.40
& 63.20
& 84.15
& 66.25
& 46.93
& 67.72
& 34.60
& 56.06
& 51.33
& 45.07 \\

\textbf{(7)}
& + L2M
& 41.38
& 2.67
& 14.47
& 35.81
& 23.58
& 50.00
& 63.80
& 84.00
& 65.93
& 46.52
& 68.52
& 34.38
& 55.97
& 51.35
& 45.23 \\

\textbf{(8)}
& + PlaM
& 36.37
& 2.67
& 14.61
& 35.69
& 22.34
& 50.80
& 62.80
& 83.32
& 65.64
& 46.47
& 67.45
& 34.60
& 55.89
& 51.10
& 44.61 \\

\midrule

\rowcolor{gray!10}
\textbf{(9)}
& \textbf{+ NeuPAT}
& \textbf{53.88}
& \textbf{10.67}
& \textbf{15.74}
& \textbf{36.39}
& \textbf{29.17}
& \textbf{56.40}
& \textbf{66.60}
& \textbf{85.97}
& \textbf{69.66}
& 50.26
& 69.53
& \textbf{37.27}
& 56.91
& \textbf{53.49}
& \textbf{49.06} \\

\rowcolor{gray!10}
\multicolumn{2}{c|}{$\boldsymbol{\Delta(9)\!-\!(2)}$}
& \pos{+16.76}
& \pos{+7.99}
& \pos{+5.84}
& \pos{+3.44}
& \pos{\textbf{+8.51}}
& \pos{+6.00}
& \pos{+4.00}
& \pos{+2.50}
& \pos{\textbf{+4.17}}
& \pos{+3.48}
& \pos{+2.16}
& \pos{+3.12}
& \pos{+1.19}
& \pos{\textbf{+2.49}}
& \pos{\textbf{+5.14}} \\

\bottomrule
\end{tabular}
\end{adjustbox}
\end{table*}

\subsection{Role-Aware Multimodal Tuning}

Based on the assigned adaptation roles, NeuPAT regulates neuron-wise update flexibility during multimodal instruction tuning.

Language-associated neurons are frozen to preserve pretrained language representations. 
Multimodal-adaptive neurons receive full updates to absorb new multimodal knowledge. 
Reserve neurons are also fully optimized as additional adaptation capacity. 
Shared neurons, which participate in both language and multimodal processing, receive constrained updates to balance preservation and adaptation.

For shared neurons, we separate the corresponding parameters into input- and output-side components, denoted by $W^{\mathrm{in}}_{l,u}$ and $W^{\mathrm{out}}_{l,u}$. 
Their deviation from pretrained parameters is constrained by:
\begingroup
\setlength{\abovedisplayskip}{6pt}
\setlength{\abovedisplayshortskip}{6pt}
\setlength{\belowdisplayskip}{6pt}
\setlength{\belowdisplayshortskip}{6pt}
\begin{equation}
\begin{aligned}
&\mathcal{R}_{\mathrm{shared}}
=
\sum_l
\sum_{u\in\mathcal{C}^{\mathrm{shared}}_l}
\Bigg[
\lambda_{\mathrm{in}}
\left\|
W^{\mathrm{in}}_{l,u}
-
W^{\mathrm{in},0}_{l,u}
\right\|_F^2
\\
&\qquad
+
\lambda_{\mathrm{out}}
\left(
1-
\cos
(
W^{\mathrm{out}}_{l,u},
W^{\mathrm{out},0}_{l,u}
)
\right)
\Bigg].
\end{aligned}
\end{equation}
\endgroup
The input-side constraint limits changes to neuron activation behavior, while the output-side cosine constraint preserves the direction of neuron contributions to the residual stream. 
During optimization, gradients of language-associated neurons are masked, while other neuron roles follow their assigned update constraints.

The final training objective is
$\mathcal{L}
=
\mathcal{L}_{\mathrm{ori}}
+
\mathcal{R}_{\mathrm{shared}},
$where $\mathcal{L}_{\mathrm{ori}}$ denotes the original autoregressive language-modeling objective.

\begin{table}[t]
\centering
\caption{
Comparison on multimodal benchmarks.
$\Delta$ denotes NeuPAT $-$ Vanilla Tuning.
}
\label{tab:multimodal_results}

\renewcommand{\arraystretch}{1.10}
\setlength{\tabcolsep}{3.7pt}
\fontsize{8pt}{8pt}\selectfont

\begin{adjustbox}{max width=\columnwidth}
\begin{tabular}{
c|
ccccc|
>{\columncolor{gray!5}}c
}
\toprule

\multicolumn{1}{c|}{\textbf{Method}}
& \shortstack{\textbf{MMB}}
& \shortstack{\textbf{RWQA}}
& \textbf{MMMU}
& \textbf{POPE}
& \textbf{MMS}
& \textbf{Avg.} \\

\midrule

Vanilla Tuning
& 72.08
& 56.60
& 43.11
& 87.69
& 45.40
& 60.98 \\

\midrule

LoRA
& 68.21
& 53.59
& 39.44
& 86.57
& 42.56
& 58.07 \\

EWC
& 67.53
& 50.59
& 43.78
& 84.99
& 40.67
& 57.51 \\

WINGS
& 73.88
& 57.78
& 43.44
& 87.86
& 47.91
& 62.17 \\

TIES
& 72.42
& 55.56
& 43.00
& 87.74
& 45.81
& 60.91 \\

L2M
& 72.94
& 55.16
& 42.89
& 87.67
& 46.24
& 60.98 \\

PlaM
& 72.77
& 56.47
& 43.00
& 87.77
& 45.10
& 61.02 \\

\midrule

\rowcolor{gray!10}
\textbf{NeuPAT}
& 72.48
& 56.99
& 42.22
& 88.72
& 44.85
& 61.05 \\

\rowcolor{gray!10}
\multicolumn{1}{c|}{$\boldsymbol{\Delta}$}
& \pos{+0.40}
& \pos{+0.39}
& \negv{-0.89}
& \pos{+1.03}
& \negv{-0.55}
& \pos{+0.07} \\

\bottomrule
\end{tabular}
\end{adjustbox}
\end{table}

\section{Experiments}
\subsection{Experimental Settings}
\subsubsection{Implementation Details}

We use RICE-ViT-Large-Patch14-560 as the vision tower~\cite{xie2025region} and Qwen3-4B-Instruct-2507 as the default language backbone~\cite{yang2025qwen3}. Generalization is evaluated on five additional backbones: Qwen3-0.6B, Phi-4-Mini-Instruct, Qwen2.5-7B-Instruct, Llama3.1-8B-Instruct, and Qwen2.5-14B-Instruct~\cite{yang2025qwen3,abouelenin2025phi,qwen25technicalreport,grattafiori2024llama}. Training consists of image-text alignment on LLaVA-558K~\cite{liu2024improved} with only the multimodal adapter updated, followed by NeuPAT-based visual instruction tuning on LLaVA-NeXT-780K~\cite{liu2024llavanext}. We set $\tau_a=0.8$ and $\lambda_{\mathrm{in}}=\lambda_{\mathrm{out}}=0.1$. All models are trained with Adam on 8 NVIDIA A100 GPUs.

\subsubsection{Baselines}

We compare NeuPAT with three baseline groups: 
\textbf{(1) reference models,} including the original LLM and Vanilla Tuning; 
\textbf{(2) general adaptation methods,} including LoRA~\cite{hu2022lora} and EWC~\cite{kirkpatrick2017overcoming}; 
and \textbf{(3) MLLM-specific preservation methods,} including the architecture-based WINGS~\cite{zhang2024wings} and post-hoc merging approaches such as TIES~\cite{ratzlaff2024training}, Locate-then-Merge~\cite{yu2025locate}, and PlaM~\cite{wang2026plam}. The original LLM provides a reference for language capability, while Vanilla Tuning reflects the forgetting caused by standard visual instruction tuning.

\subsubsection{Evaluation Protocol}

We evaluate language and multimodal capabilities using \texttt{lm-evaluation-harness}~\cite{eval-harness} and \texttt{lmms-eval}~\cite{zhang2025lmms}, respectively. The language suite contains 11 benchmarks covering knowledge, mathematics, logic, coding, and reading comprehension: SocialIQA~\cite{sap2019social}, ARC-Challenge~\cite{arc}, GPQA~\cite{gpqa}, MATH-500~\cite{math}, GSM8K~\cite{gsm8k}, LogiQA2~\cite{logiqa2}, BBH~\cite{bbh}, bAbI~\cite{babi}, MBPP~\cite{mbpp}, DROP~\cite{drop} and CoQA~\cite{coqa}. Multimodal performance is evaluated on MMBench-EN~\cite{mmbench}, RealWorldQA~\cite{realworldqa}, POPE~\cite{pope}, MMMU~\cite{mmmu} and MMStar~\cite{mmstar}.

\begin{table*}[t]
\centering
\caption{
Generalization across different LLM backbones. 
For each backbone, we report the original LLM, Vanilla Tuning, and our method. 
We show representative language benchmarks and multimodal benchmarks.
$\Delta$ denotes NeuPAT $-$ Vanilla Tuning.
}
\label{tab:backbone_generalization}

\renewcommand{\arraystretch}{1.10}
\setlength{\tabcolsep}{3.0pt}
\fontsize{7pt}{7pt}\selectfont

\begin{tabular}{l|l|ccccc|c|ccccc|c}
\toprule

\multirow{2}{*}{\textbf{Backbone}} 
& \multirow{2}{*}{\textbf{Method}}
& \multicolumn{6}{c|}{\textbf{Language Benchmarks}}
& \multicolumn{6}{c}{\textbf{Multimodal Benchmarks}} \\

\cmidrule{3-8}
\cmidrule{9-14}

&
& \textbf{SIQA}
& \textbf{GSM8K}
& \textbf{BBH}
& \textbf{MBPP}
& \textbf{CoQA}
& \textbf{Avg.}
& \textbf{MMB}
& \textbf{RWQA}
& \textbf{MMMU}
& \textbf{POPE}
& \textbf{MMStar}
& \textbf{Avg.} \\

\midrule

& LLM
& 40.38 & 40.56 & 33.39 & 27.40 & 57.57
& 39.86
& -- & -- & -- & -- & --
& -- \\

& + Vanilla Tuning
& 39.00 & 35.18 & 28.69 & 22.60 & 55.93
& 36.28
& 52.58 & 45.10 & 30.67 & 85.42 & 37.23
& 50.20 \\

\rowcolor{gray!9}
\cellcolor{white}
& \textbf{+ NeuPAT}
& 40.33 & 40.11 & 33.44 & 28.20 & 59.42
& 40.30
& 52.52 & 44.75 & 32.67 & 84.26 & 37.07
& 50.25 \\

\rowcolor{gray!9}
\cellcolor{white}
\multirow{-4}{*}{Qwen3-0.6B}
& \multicolumn{1}{c|}{$\boldsymbol{\Delta}$}
& \pos{+1.33} & \pos{+4.93} & \pos{+4.75} & \pos{+5.60} & \pos{+3.49}
& \pos{+4.02}
& \negv{-0.06} & \negv{-0.35} & \pos{+2.00} & \negv{-1.16} & \negv{-0.16}
& \pos{+0.05} \\

\midrule

& LLM
& 49.59 & 83.62 & 52.82 & 55.40 & 78.40
& 63.97
& -- & -- & -- & -- & --
& -- \\

& + Vanilla Tuning
& 46.11 & 73.92 & 40.04 & 47.60 & 65.23
& 54.58
& 50.52 & 45.10 & 36.78 & 77.16 & 30.58
& 48.03 \\

\rowcolor{gray!9}
\cellcolor{white}
& \textbf{+ NeuPAT}
& 50.06 & 81.45 & 51.29 & 53.80 & 80.04
& 63.33
& 50.89 & 44.31 & 36.22 & 77.82 & 31.26
& 48.10 \\

\rowcolor{gray!9}
\cellcolor{white}
\multirow{-4}{*}{Phi-4-Mini-Instruct}
& \multicolumn{1}{c|}{$\boldsymbol{\Delta}$}
& \pos{+3.95} & \pos{+7.53} & \pos{+11.25} & \pos{+6.20} & \pos{+14.81}
& \pos{+8.75}
& \pos{+0.37} & \negv{-0.79} & \negv{-0.56} & \pos{+0.66} & \pos{+0.68}
& \pos{+0.07} \\

\midrule

& LLM
& 51.59 & 76.50 & 45.81 & 47.60 & 78.74
& 60.05
& -- & -- & -- & -- & --
& -- \\

& + Vanilla Tuning
& 50.20 & 74.83 & 41.98 & 40.20 & 73.63
& 56.17
& 64.78 & 52.03 & 41.00 & 87.12 & 40.56
& 57.10 \\

\rowcolor{gray!9}
\cellcolor{white}
& \textbf{+ NeuPAT}
& 55.89 & 78.70 & 47.53 & 42.80 & 78.80
& 60.74
& 64.64 & 52.01 & 42.56 & 86.54 & 40.37
& 57.22 \\

\rowcolor{gray!9}
\cellcolor{white}
\multirow{-4}{*}{Qwen2.5-7B-Instruct}
& \multicolumn{1}{c|}{$\boldsymbol{\Delta}$}
& \pos{+5.69} & \pos{+3.87} & \pos{+5.55} & \pos{+2.60} & \pos{+5.17}
& \pos{+4.57}
& \negv{-0.14} & \negv{-0.02} & \pos{+1.56} & \negv{-0.58} & \negv{-0.19}
& \pos{+0.12} \\

\midrule

& LLM
& 49.85 & 78.17 & 44.59 & 58.40 & 77.83
& 61.77
& -- & -- & -- & -- & --
& -- \\

& + Vanilla Tuning
& 49.74 & 67.55 & 34.91 & 54.20 & 77.50
& 56.78
& 60.14 & 26.67 & 35.89 & 82.23 & 36.54
& 48.29 \\

\rowcolor{gray!9}
\cellcolor{white}
& \textbf{+ NeuPAT}
& 49.95 & 74.55 & 41.68 & 56.80 & 80.48
& 60.69
& 61.94 & 44.97 & 36.22 & 80.90 & 33.94
& 51.59 \\

\rowcolor{gray!9}
\cellcolor{white}
\multirow{-4}{*}{Llama3.1-8B-Instruct}
& \multicolumn{1}{c|}{$\boldsymbol{\Delta}$}
& \pos{+0.21} & \pos{+7.00} & \pos{+6.77} & \pos{+2.60} & \pos{+2.98}
& \pos{+3.91}
& \pos{+1.80} & \pos{+18.30} & \pos{+0.33} & \negv{-1.33} & \negv{-2.60}
& \pos{+3.30} \\

\midrule

& LLM
& 54.04 & 79.83 & 52.54 & 66.80 & 78.20
& 66.28
& -- & -- & -- & -- & --
& -- \\

& + Vanilla Tuning
& 51.23 & 79.30 & 39.87 & 65.60 & 74.44
& 62.09
& 75.17 & 56.34 & 47.67 & 87.27 & 50.26
& 63.34 \\

\rowcolor{gray!9}
\cellcolor{white}
& \textbf{+ NeuPAT}
& 55.22 & 85.44 & 52.59 & 66.80 & 77.51
& 67.51
& 74.31 & 55.64 & 48.11 & 87.67 & 50.14
& 63.17 \\

\rowcolor{gray!9}
\cellcolor{white}
\multirow{-4}{*}{Qwen2.5-14B-Instruct}
& \multicolumn{1}{c|}{$\boldsymbol{\Delta}$}
& \pos{+3.99} & \pos{+6.14} & \pos{+12.72} & \pos{+1.20} & \pos{+3.07}
& \pos{+5.42}
& \negv{-0.86} & \negv{-0.70} & \pos{+0.44} & \pos{+0.40} & \negv{-0.12}
& \negv{-0.17} \\

\bottomrule
\end{tabular}
\end{table*}

\subsection{Experimental Results}
\subsubsection{Main Results}

Tables~\ref{tab:language_results} and~\ref{tab:multimodal_results} report results on 11 language and 5 multimodal benchmarks. Vanilla Tuning lowers the language average from $48.99$ to $43.92$, with drops exceeding 10 points on BBH, bAbI, and MATH-500. NeuPAT recovers $5.14$ points, reaching $49.06$ and slightly surpassing the original LLM, with particularly strong gains on language and logical reasoning tasks. Meanwhile, it maintains comparable multimodal performance, improving the average from $60.98$ to $61.05$. Although WINGS achieves the highest multimodal average, NeuPAT provides the best overall language performance and a stronger balance between language preservation and multimodal adaptation.

\subsubsection{Generalization across LLM Backbones}

To evaluate cross-backbone generalization, we replace the default LLM with Qwen3-0.6B, Phi-4-Mini-Instruct, Qwen2.5-7B/14B-Instruct, and Llama3.1-8B-Instruct. Due to space limitations, Table~\ref{tab:backbone_generalization} reports five representative language benchmarks, while the complete results are provided in the appendix. Vanilla Tuning consistently degrades language performance across all tested backbones, whereas NeuPAT improves the reported text average over Vanilla Tuning by $4.02$, $8.75$, $4.57$, $3.91$, and $5.42$ points, respectively, while maintaining comparable multimodal performance. For Qwen3-0.6B and both Qwen2.5 backbones, NeuPAT even surpasses the original LLM average. These consistent improvements across different model families and scales demonstrate that NeuPAT is not tied to a specific language backbone.

\begin{table*}[t] 
\centering
\caption{Ablation of neuron-wise adaptation role allocation.
\checkmark: full update, $\times$: freeze, Reg.: regularized update.}
\label{tab:ablation_c}

\renewcommand{\arraystretch}{1.10} 
\setlength{\tabcolsep}{4.0pt}     
\fontsize{7pt}{7pt}\selectfont    

\begin{tabular}{c|cccc|ccc|c|ccc|c}
\toprule

\multirow{2}{*}{\textbf{Variants}}
& \multicolumn{4}{c|}{\textbf{Neuron Adaptation Role}}
& \multicolumn{4}{c|}{\textbf{Language Benchmarks}}
& \multicolumn{4}{c}{\textbf{Multimodal Benchmarks}} \\

\cmidrule(lr){2-5}
\cmidrule(lr){6-9}
\cmidrule(lr){10-13}

& \textbf{Lang.}
& \textbf{Multi.}
& \textbf{Shared}
& \textbf{Reserve}
& \textbf{GSM8K}
& \textbf{MBPP}
& \textbf{bAbI}
& \textbf{Avg.}
& \textbf{MMB}
& \textbf{RWQA}
& \textbf{POPE}
& \textbf{Avg.} \\

\midrule

w/o Language Freeze
& \checkmark & \checkmark & Reg. & \checkmark
& 83.90 & 64.80 & 5.87 & 51.52
& 72.11 & 57.30 & 87.20 & 72.20 \\

w/o Multimodal Update
& $\times$ & $\times$ & Reg. & \checkmark
& 84.44 & 67.10 & 11.85 & 54.46
& 69.69 & 55.16 & 86.61 & 70.49 \\

w/o Reserve Update
& $\times$ & \checkmark & Reg. & $\times$
& 84.76 & 66.80 & 11.64 & 54.40
& 71.05 & 56.34 & 86.94 & 71.44 \\

Shared Full Update
& $\times$ & \checkmark & \checkmark & \checkmark
& 83.61 & 64.00 & 9.94 & 52.52
& 72.34 & 57.25 & 87.27 & 72.29 \\

Shared Freeze
& $\times$ & \checkmark & $\times$ & \checkmark
& 85.06 & 67.20 & 15.16 & 55.81
& 69.93 & 55.03 & 87.30 & 70.75 \\

\rowcolor{gray!10}
\textbf{Ours}
& $\times$ & \checkmark & Reg. & \checkmark
& 85.97 & 66.60 & 10.67 & 54.41
& 72.48 & 56.99 & 88.72 & 72.73 \\

\bottomrule
\end{tabular}
\end{table*}

\begin{table}[t]
\centering
\caption{Ablation of global update schemes and role-aware update allocation.}
\label{tab:ablation_a}

\renewcommand{\arraystretch}{1.10}
\setlength{\tabcolsep}{2.0pt}
\scriptsize

\begin{tabular}{c|ccc|c|ccc|c}
\toprule
\textbf{Update}
& \multicolumn{4}{c|}{\textbf{Language Benchmarks}}
& \multicolumn{4}{c}{\textbf{Multimodal Benchmarks}} \\
\cmidrule(lr){2-5}
\cmidrule(lr){6-9}
\textbf{strategy}
& \textbf{GSM8K}
& \textbf{MBPP}
& \textbf{bAbI}
& \textbf{Avg.}
& \textbf{MMB}
& \textbf{RWQA}
& \textbf{POPE}
& \textbf{Avg.} \\

\midrule

Freeze
& 86.05 & 67.10 & 13.65 & 55.60
& 68.64 & 53.99 & 86.03 & 69.55 \\

Update
& 83.47 & 62.60 & 2.68 & 49.58
& 72.08 & 56.60 & 87.69 & 72.12 \\

Reg.
& 84.99 & 65.50 & 9.33 & 53.27
& 72.08 & 54.25 & 86.03 & 70.79 \\

\rowcolor{gray!10}
\textbf{Ours}
& 85.97 & 66.60 & 10.67 & 54.41
& 72.48 & 56.99 & 88.72 & 72.73 \\

\bottomrule
\end{tabular}
\end{table}

\begin{table}[t]
\centering
\caption{Ablation of neuron role allocation strategies.}
\label{tab:ablation_b}

\renewcommand{\arraystretch}{1.10}
\setlength{\tabcolsep}{2.0pt}
\scriptsize

\begin{tabular}{c|ccc|c|ccc|c}
\toprule

\textbf{Allocation}
& \multicolumn{4}{c|}{\textbf{Language Benchmarks}}
& \multicolumn{4}{c}{\textbf{Multimodal Benchmarks}} \\

\cmidrule(lr){2-5}
\cmidrule(lr){6-9}
\textbf{strategy}
& \textbf{GSM8K}
& \textbf{MBPP}
& \textbf{bAbI}
& \textbf{Avg.}
& \textbf{MMB}
& \textbf{RWQA}
& \textbf{POPE}
& \textbf{Avg.} \\

\midrule

Random
& 83.69 & 65.40 & 1.94 & 50.34
& 69.76 & 54.51 & 87.52 & 70.60 \\

Fixed-ratio
& 84.90 & 66.20 & 10.24 & 53.78
& 72.16 & 57.25 & 87.28 & 72.23 \\

\rowcolor{gray!10}
\textbf{Ours}
& 85.97 & 66.60 & 10.67 & 54.41
& 72.48 & 56.99 & 88.72 & 72.73 \\

\bottomrule
\end{tabular}
\end{table}

\subsubsection{Ablation Studies}

\paragraph{Global vs. Neuron-Aware Update.}
We compare NeuPAT with three global update schemes. 
\textit{Freeze} freezes all neurons, \textit{Update} uniformly updates the backbone, and \textit{Reg.} applies the shared-neuron regularization globally. 
As shown in Table~\ref{tab:ablation_a}, global strategies suffer from either limited multimodal adaptation or insufficient language preservation. 
NeuPAT achieves the best balance, obtaining language and multimodal averages of $54.41$ and $72.73$, demonstrating the effectiveness of neuron-wise update allocation.

\paragraph{Neuron Allocation Strategy.}
We compare importance-guided role allocation with \textit{Random} and \textit{Fixed-ratio} baselines. 
\textit{Random} preserves category sizes but randomly assigns neuron roles, while \textit{Fixed-ratio} applies the same selection ratio across layers. 
As shown in Table~\ref{tab:ablation_b}, NeuPAT consistently outperforms both baselines, validating the importance of reliable neuron identification and layer-adaptive allocation.

\paragraph{Neuron-Wise Update Constraint.}
We further evaluate each role-specific update strategy by modifying one constraint at a time. 
As shown in Table~\ref{tab:ablation_c}, updating language-associated neurons decreases text performance, while freezing vision-associated or reserve neurons limits multimodal adaptation. 
For shared neurons, full updating harms language preservation and full freezing sacrifices multimodal performance. 
These results support the proposed strategy of freezing language neurons, updating multimodal and reserve neurons, and regularizing shared neurons.

\subsubsection{Neuron Distribution Visualization} 
\begin{figure}[t]
\centering
\includegraphics[width=0.48\textwidth]{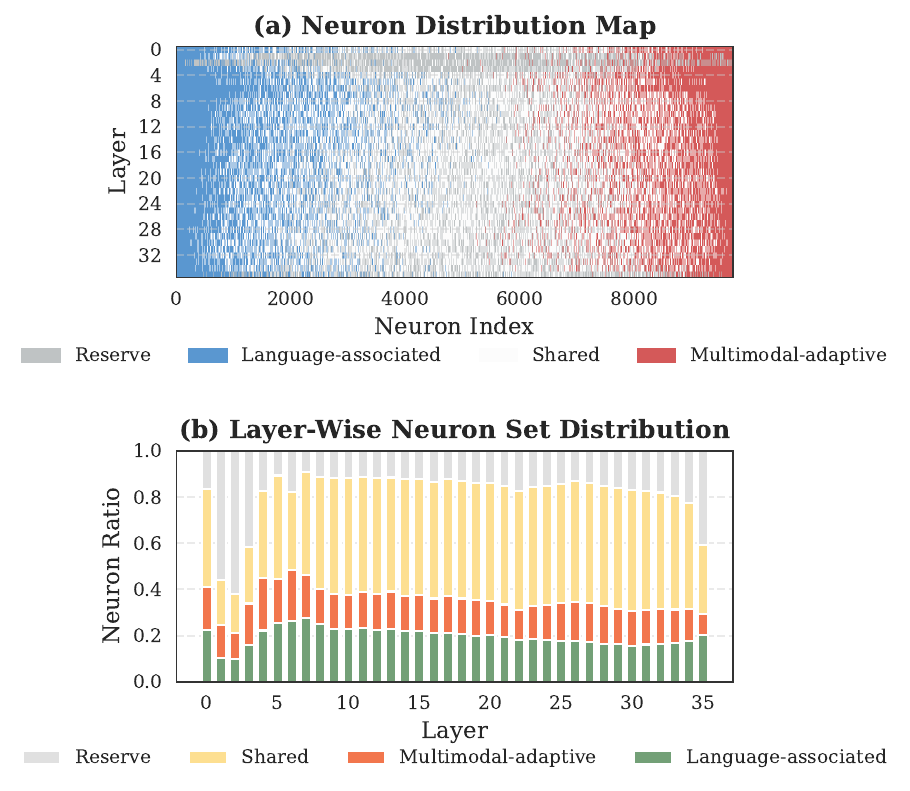} 
\caption{\textbf{Layer-wise adaptation role distribution.}
(a) Adaptation roles identified by importance-guided allocation.
(b) Layer-wise proportions of neuron roles.}
\label{fig:neuron_distribution}
\end{figure}

We visualize the assigned neuron roles and their layer-wise distributions in Figure~\ref{fig:neuron_distribution}. As shown in Figure~\ref{fig:neuron_distribution}(a), the resulting neuron map is closely consistent with the modality-associated response patterns observed in Figure~\ref{motivation}. Language-associated neurons are concentrated in text-preferred regions, whereas multimodal-adaptive neurons primarily occupy vision-preferred regions. In contrast, shared and reserve neurons are located mainly in balanced-response regions, but differ markedly in their overall response strength: shared neurons respond strongly to both input types, while reserve neurons remain weakly engaged and may provide underutilized capacity for multimodal adaptation. Figure~\ref{fig:neuron_distribution}(b) further shows that shared neurons constitute the largest group in most layers, accounting for approximately $45\%$, while language-associated and multimodal-adaptive neurons each represent around $20\%$, and reserve neurons account for roughly $15\%$. These consistent yet layer-dependent distributions reveal substantial heterogeneity in neuron functionality and adaptation capacity, supporting the need for layer-adaptive neuron allocation and differentiated plasticity control within the LLM backbone.

\section{Conclusion}
Multimodal expansion enables LLMs to acquire capabilities beyond language, but may compromise the language intelligence inherited from pretraining. We show that this degradation stems from heterogeneous adaptation behaviors within the pretrained LLM backbone and introduce NeuPAT, a neuron-aware update allocation framework for capability-preserving multimodal tuning. By selectively regulating neuron update flexibility, NeuPAT balances language preservation and multimodal adaptation without additional data, architectural changes, or post-training merging. Experiments across diverse LLM backbones demonstrate that NeuPAT is an efficient, architecture-agnostic solution for reliable LLM-to-MLLM expansion.

\section{Limitations}
NeuPAT has several limitations. 
First, although we evaluate it across diverse LLM families and model scales, its effectiveness on substantially larger backbones remains unverified. 
Larger models may exhibit more distributed and complex adaptation patterns, and the current neuron-wise role allocation mechanism may require further investigation at greater scale.
Second, our experiments focus on vision-language expansion. 
Extending NeuPAT to other modalities, such as audio, video, embodied interaction, or unified multimodal systems, may introduce different alignment dynamics and modality interactions, potentially requiring modality-specific importance estimation and update strategies.
Finally, this work considers a single-stage multimodal expansion process. 
In continual or sequential learning scenarios, newly introduced modalities and tasks may interact with both pretrained language capabilities and previously acquired multimodal knowledge. 
Developing mechanisms for stable long-term capability accumulation without progressive interference remains an important direction for future research.

\section{Ethical Considerations}
All training data and evaluation experiments in this work are based on publicly available datasets and benchmarks. And we emphasize that the proposed neuron allocation strategies are response-based functional approximations rather than causal explanations of model behavior. Models trained with NeuPAT should therefore undergo standard safety, fairness, and robustness evaluations before deployment, especially in high-stakes applications. AI tools were used only for language polishing. All research ideas, experiments, analyses, and manuscript organization were completed by the authors.

\bibliography{custom}

@article{liu2023visual,
  title={Visual instruction tuning},
  author={Liu, Haotian and Li, Chunyuan and Wu, Qingyang and Lee, Yong Jae},
  journal={Advances in neural information processing systems},
  volume={36},
  pages={34892--34916},
  year={2023}
}

@article{dai2023instructblip,
  title={Instructblip: Towards general-purpose vision-language models with instruction tuning},
  author={Dai, Wenliang and Li, Junnan and Li, Dongxu and Tiong, Anthony and Zhao, Junqi and Wang, Weisheng and Li, Boyang and Fung, Pascale N and Hoi, Steven},
  journal={Advances in neural information processing systems},
  volume={36},
  pages={49250--49267},
  year={2023}
}

@article{wang2024qwen2,
  title={Qwen2-vl: Enhancing vision-language model's perception of the world at any resolution},
  author={Wang, Peng and Bai, Shuai and Tan, Sinan and Wang, Shijie and Fan, Zhihao and Bai, Jinze and Chen, Keqin and Liu, Xuejing and Wang, Jialin and Ge, Wenbin and others},
  journal={arXiv preprint arXiv:2409.12191},
  year={2024}
}

@article{bai2025qwen3,
  title={Qwen3-vl technical report},
  author={Bai, Shuai and Cai, Yuxuan and Chen, Ruizhe and Chen, Keqin and Chen, Xionghui and Cheng, Zesen and Deng, Lianghao and Ding, Wei and Gao, Chang and Ge, Chunjiang and others},
  journal={arXiv preprint arXiv:2511.21631},
  year={2025}
}

@article{li2024llava,
  title={Llava-onevision: Easy visual task transfer},
  author={Li, Bo and Zhang, Yuanhan and Guo, Dong and Zhang, Renrui and Li, Feng and Zhang, Hao and Zhang, Kaichen and Zhang, Peiyuan and Li, Yanwei and Liu, Ziwei and others},
  journal={arXiv preprint arXiv:2408.03326},
  year={2024}
}

@article{li2026mnaft,
  title={MNAFT: modality neuron-aware fine-tuning of multimodal large language models for image translation},
  author={Li, Bo and Deng, Ningyuan and Dong, Tianyu and Wang, Shaobo and Zhu, Shaolin and Wen, Lijie},
  journal={Science China Information Sciences},
  volume={69},
  number={5},
  year={2026},
  publisher={Science China Press}
}

@article{zhu2025internvl3,
  title={Internvl3: Exploring advanced training and test-time recipes for open-source multimodal models},
  author={Zhu, Jinguo and Wang, Weiyun and Chen, Zhe and Liu, Zhaoyang and Ye, Shenglong and Gu, Lixin and Tian, Hao and Duan, Yuchen and Su, Weijie and Shao, Jie and others},
  journal={arXiv preprint arXiv:2504.10479},
  year={2025}
}

@article{team2025kimi,
  title={Kimi-vl technical report},
  author={Team, Kimi and Du, Angang and Yin, Bohong and Xing, Bowei and Qu, Bowen and Wang, Bowen and Chen, Cheng and Zhang, Chenlin and Du, Chenzhuang and Wei, Chu and others},
  journal={arXiv preprint arXiv:2504.07491},
  year={2025}
}

@article{wang2025internvl3,
  title={Internvl3. 5: Advancing open-source multimodal models in versatility, reasoning, and efficiency},
  author={Wang, Weiyun and Gao, Zhangwei and Gu, Lixin and Pu, Hengjun and Cui, Long and Wei, Xingguang and Liu, Zhaoyang and Jing, Linglin and Ye, Shenglong and Shao, Jie and others},
  journal={arXiv preprint arXiv:2508.18265},
  year={2025}
}

@article{zhang2024wings,
  title={Wings: Learning multimodal llms without text-only forgetting},
  author={Zhang, Yi-Kai and Lu, Shiyin and Li, Yang and Ma, Yanqing and Chen, Qing-Guo and Xu, Zhao and Luo, Weihua and Zhang, Kaifu and Zhan, De-Chuan and Ye, Han-Jia},
  journal={Advances in Neural Information Processing Systems},
  volume={37},
  pages={31828--31853},
  year={2024}
}

@article{ratzlaff2024training,
  title={Training-free mitigation of language reasoning degradation after multimodal instruction tuning},
  author={Ratzlaff, Neale and Luo, Man and Su, Xin and Lal, Vasudev and Howard, Phillip},
  journal={arXiv preprint arXiv:2412.03467},
  year={2024}
}

@article{lu2024deepseek,
  title={Deepseek-vl: towards real-world vision-language understanding},
  author={Lu, Haoyu and Liu, Wen and Zhang, Bo and Wang, Bingxuan and Dong, Kai and Liu, Bo and Sun, Jingxiang and Ren, Tongzheng and Li, Zhuoshu and Yang, Hao and others},
  journal={arXiv preprint arXiv:2403.05525},
  year={2024}
}

@article{hu2022lora,
  title={Lora: Low-rank adaptation of large language models.},
  author={Hu, Edward J and Shen, Yelong and Wallis, Phillip and Allen-Zhu, Zeyuan and Li, Yuanzhi and Wang, Shean and Wang, Liang and Chen, Weizhu and others},
  journal={Iclr},
  volume={1},
  number={2},
  pages={3},
  year={2022}
}

@article{brown2020language,
  title={Language models are few-shot learners},
  author={Brown, Tom and Mann, Benjamin and Ryder, Nick and Subbiah, Melanie and Kaplan, Jared D and Dhariwal, Prafulla and Neelakantan, Arvind and Shyam, Pranav and Sastry, Girish and Askell, Amanda and others},
  journal={Advances in neural information processing systems},
  volume={33},
  pages={1877--1901},
  year={2020}
}

@article{chowdhery2023palm,
  title={Palm: Scaling language modeling with pathways},
  author={Chowdhery, Aakanksha and Narang, Sharan and Devlin, Jacob and Bosma, Maarten and Mishra, Gaurav and Roberts, Adam and Barham, Paul and Chung, Hyung Won and Sutton, Charles and Gehrmann, Sebastian and others},
  journal={Journal of machine learning research},
  volume={24},
  number={240},
  pages={1--113},
  year={2023}
}

@article{openai2023gpt4,
  title={Gpt-4 technical report},
  author={Achiam, Josh and Adler, Steven and Agarwal, Sandhini and Ahmad, Lama and Akkaya, Ilge and Aleman, Florencia Leoni and Almeida, Diogo and Altenschmidt, Janko and Altman, Sam and Anadkat, Shyamal and others},
  journal={arXiv preprint arXiv:2303.08774},
  year={2023}
}

@article{touvron2023llama,
  title={Llama: Open and efficient foundation language models},
  author={Touvron, Hugo and Lavril, Thibaut and Izacard, Gautier and Martinet, Xavier and Lachaux, Marie-Anne and Lacroix, Timoth{\'e}e and Rozi{\`e}re, Baptiste and Goyal, Naman and Hambro, Eric and Azhar, Faisal and others},
  journal={arXiv preprint arXiv:2302.13971},
  year={2023}
}

@article{kirkpatrick2017overcoming,
  title={Overcoming catastrophic forgetting in neural networks},
  author={Kirkpatrick, James and Pascanu, Razvan and Rabinowitz, Neil and Veness, Joel and Desjardins, Guillaume and Rusu, Andrei A and Milan, Kieran and Quan, John and Ramalho, Tiago and Grabska-Barwinska, Agnieszka and others},
  journal={Proceedings of the national academy of sciences},
  volume={114},
  number={13},
  pages={3521--3526},
  year={2017},
  publisher={National Academy of Sciences}
}

@inproceedings{wang2025iaa,
  title={Iaa: Inner-adaptor architecture empowers frozen large language model with multimodal capabilities},
  author={Wang, Bin and Xie, Chunyu and Leng, Dawei and Yin, Yuhui},
  booktitle={Proceedings of the AAAI Conference on Artificial Intelligence},
  volume={39},
  pages={21035--21043},
  year={2025}
}

@inproceedings{lu2025genieblue,
  title={GenieBlue: Integrating both Linguistic and Multimodal Capabilities for Large Language Models on Mobile Devices},
  author={Lu, Xudong and Chen, Yinghao and Wu, Renshou and Gao, Haohao and Chen, Xi and Yang, Xue and Zhao, Xiangyu and Zhou, Aojun and Li, Fangyuan and Wen, Yafei and others},
  booktitle={Proceedings of the IEEE/CVF International Conference on Computer Vision},
  pages={4198--4210},
  year={2025}
}

@article{yu2025locate,
  title={Locate-then-Merge: Neuron-Level Parameter Fusion for Mitigating Catastrophic Forgetting in Multimodal LLMs},
  author={Yu, Zeping and Ananiadou, Sophia},
  journal={arXiv preprint arXiv:2505.16703},
  year={2025}
}

@article{wang2026plam,
  title={PlaM: Training-Free Plateau-Guided Model Merging for Better Visual Grounding in MLLMs},
  author={Wang, Zijing and Liu, Yongkang and Wang, Mingyang and Nie, Ercong and Chen, Deyuan and Zhao, Zhengjie and Feng, Shi and Wang, Daling and Yang, Xiaocui and Zhang, Yifei and others},
  journal={arXiv preprint arXiv:2601.07645},
  year={2026}
}

@inproceedings{li2025dpimerge,
  title={DPIMerge: An Efficient Dynamic Parameter Interpolation Framework for Alleviating Pure Text Forgetting in Multimodal Large Models},
  author={Li, Zhihua and Wen, Haoguang and Hu, Wenpeng and Luo, Zhunchen and Chen, Lingqiang and Wang, Sijia and Wu, Shuyi},
  booktitle={CCF International Conference on Natural Language Processing and Chinese Computing},
  pages={3--15},
  year={2025},
  organization={Springer}
}

@article{alayrac2022flamingo,
  title={Flamingo: a visual language model for few-shot learning},
  author={Alayrac, Jean-Baptiste and Donahue, Jeff and Luc, Pauline and Miech, Antoine and Barr, Iain and Hasson, Yana and Lenc, Karel and Mensch, Arthur and Millican, Katie and Reynolds, Malcolm and others},
  journal={arXiv preprint arXiv:2204.14198},
  year={2022}
}

@inproceedings{li2023blip,
  title={Blip-2: Bootstrapping language-image pre-training with frozen image encoders and large language models},
  author={Li, Junnan and Li, Dongxu and Savarese, Silvio and Hoi, Steven},
  booktitle={International conference on machine learning},
  pages={19730--19742},
  year={2023},
  organization={PMLR}
}

@inproceedings{zhu2024minigpt,
  title={Minigpt-4: Enhancing vision-language understanding with advanced large language models},
  author={Zhu, Deyao and Shen, Xiaoqian and Li, Xiang and Elhoseiny, Mohamed and others},
  booktitle={International Conference on Learning Representations},
  volume={2024},
  pages={18378--18394},
  year={2024}
}

@inproceedings{liu2024improved,
  title={Improved baselines with visual instruction tuning},
  author={Liu, Haotian and Li, Chunyuan and Li, Yuheng and Lee, Yong Jae},
  booktitle={Proceedings of the IEEE/CVF conference on computer vision and pattern recognition},
  pages={26296--26306},
  year={2024}
}

@article{bai2023qwen,
  title={Qwen technical report},
  author={Bai, Jinze and Bai, Shuai and Chu, Yunfei and Cui, Zeyu and Dang, Kai and Deng, Xiaodong and Fan, Yang and Ge, Wenbin and Han, Yu and Huang, Fei and others},
  journal={arXiv preprint arXiv:2309.16609},
  year={2023}
}

@article{fu2026correct,
  title={Correct When Paired, Wrong When Split: Decoupling and Editing Modality-Specific Neurons in MLLMs},
  author={Fu, Tingchao and Wang, Wenkai and Li, Fanxiao and Zhang, Huadong and Zhang, Jinhong and Li, Dayang and Dong, Yunyun and Liu, Renyang and Zhou, Wei},
  journal={arXiv preprint arXiv:2606.17057},
  year={2026}
}

@article{zhao2026discovering,
  title={Discovering and Causally Validating Emotion-Sensitive Neurons in Large Audio-Language Models},
  author={Zhao, Xiutian and Schuller, Bj{\"o}rn and Sisman, Berrak},
  journal={arXiv preprint arXiv:2601.03115},
  year={2026}
}

@inproceedings{liu2025modality,
  title={Modality-aware neuron pruning for unlearning in multimodal large language models},
  author={Liu, Zheyuan and Dou, Guangyao and Yuan, Xiangchi and Zhang, Chunhui and Tan, Zhaoxuan and Jiang, Meng},
  booktitle={Proceedings of the 63rd Annual Meeting of the Association for Computational Linguistics (Volume 1: Long Papers)},
  pages={5913--5933},
  year={2025}
}

@article{qin2025achilles,
  title={The Achilles' Heel of LLMs: How Altering a Handful of Neurons Can Cripple Language Abilities},
  author={Qin, Zixuan and Yu, Qingchen and Lyu, Kunlin and Fan, Zhaoxin and Sun, Yifan},
  journal={arXiv preprint arXiv:2510.10238},
  year={2025}
}

@inproceedings{sun2024simple,
  title={A simple and effective pruning approach for large language models},
  author={Sun, Mingjie and Liu, Zhuang and Bair, Anna and Kolter, Zico},
  booktitle={International Conference on Learning Representations},
  volume={2024},
  pages={4942--4964},
  year={2024}
}

@inproceedings{xie2025region,
  title={Region-based cluster discrimination for visual representation learning},
  author={Xie, Yin and Yang, Kaicheng and An, Xiang and Wu, Kun and Zhao, Yongle and Deng, Weimo and Ran, Zimin and Wang, Yumeng and Feng, Ziyong and Miles, Roy and others},
  booktitle={Proceedings of the IEEE/CVF International Conference on Computer Vision},
  pages={1793--1803},
  year={2025}
}

@article{abouelenin2025phi,
  title={Phi-4-mini technical report: Compact yet powerful multimodal language models via mixture-of-loras},
  author={Abouelenin, Abdelrahman and Ashfaq, Atabak and Atkinson, Adam and Awadalla, Hany and Bach, Nguyen and Bao, Jianmin and Benhaim, Alon and Cai, Martin and Chaudhary, Vishrav and Chen, Congcong and others},
  journal={arXiv preprint arXiv:2503.01743},
  year={2025}
}

@article{yang2025qwen3,
  title={Qwen3 technical report},
  author={Yang, An and Li, Anfeng and Yang, Baosong and Zhang, Beichen and Hui, Binyuan and Zheng, Bo and Yu, Bowen and Gao, Chang and Huang, Chengen and Lv, Chenxu and others},
  journal={arXiv preprint arXiv:2505.09388},
  year={2025}
}

@article{qwen25technicalreport,
  title   = {Qwen2.5 Technical Report},
  author  = {Yang, An and Yang, Baosong and Zhang, Beichen and
             Hui, Binyuan and Zheng, Bo and Yu, Bowen and others},
  journal = {arXiv preprint arXiv:2412.15115},
  year    = {2024}
}

@article{grattafiori2024llama,
  title={The llama 3 herd of models},
  author={Grattafiori, Aaron and Dubey, Abhimanyu and Jauhri, Abhinav and Pandey, Abhinav and Kadian, Abhishek and Al-Dahle, Ahmad and Letman, Aiesha and Mathur, Akhil and Schelten, Alan and Vaughan, Alex and others},
  journal={arXiv preprint arXiv:2407.21783},
  year={2024}
}

@misc{liu2024llavanext,
    title={LLaVA-NeXT: Improved reasoning, OCR, and world knowledge},
    url={https://llava-vl.github.io/blog/2024-01-30-llava-next/},
    author={Liu, Haotian and Li, Chunyuan and Li, Yuheng and Li, Bo and Zhang, Yuanhan and Shen, Sheng and Lee, Yong Jae},
    month={January},
    year={2024}
}

@misc{eval-harness,
  author       = {Gao, Leo and Tow, Jonathan and Abbasi, Baber and Biderman, Stella and Black, Sid and DiPofi, Anthony and Foster, Charles and Golding, Laurence and Hsu, Jeffrey and Le Noac'h, Alain and Li, Haonan and McDonell, Kyle and Muennighoff, Niklas and Ociepa, Chris and Phang, Jason and Reynolds, Laria and Schoelkopf, Hailey and Skowron, Aviya and Sutawika, Lintang and Tang, Eric and Thite, Anish and Wang, Ben and Wang, Kevin and Zou, Andy},
  title        = {The Language Model Evaluation Harness},
  month        = 07,
  year         = 2024,
  publisher    = {Zenodo},
  version      = {v0.4.3},
  doi          = {10.5281/zenodo.12608602},
  url          = {https://zenodo.org/records/12608602}
}

@inproceedings{zhang2025lmms,
  title={Lmms-eval: Reality check on the evaluation of large multimodal models},
  author={Zhang, Kaichen and Li, Bo and Zhang, Peiyuan and Pu, Fanyi and Cahyono, Joshua Adrian and Hu, Kairui and Liu, Shuai and Zhang, Yuanhan and Yang, Jingkang and Li, Chunyuan and others},
  booktitle={Findings of the Association for Computational Linguistics: NAACL 2025},
  pages={881--916},
  year={2025}
}

@inproceedings{sap2019social,
  title={Social IQa: Commonsense reasoning about social interactions},
  author={Sap, Maarten and Rashkin, Hannah and Chen, Derek and Le Bras, Ronan and Choi, Yejin},
  booktitle={Proceedings of the 2019 conference on empirical methods in natural language processing and the 9th international joint conference on natural language processing (EMNLP-IJCNLP)},
  pages={4463--4473},
  year={2019}
}

@article{gsm8k,
  title={Training verifiers to solve math word problems},
  author={Cobbe, Karl and Kosaraju, Vineet and Bavarian, Mohammad and Chen, Mark and Jun, Heewoo and Kaiser, Lukasz and Plappert, Matthias and Tworek, Jerry and Hilton, Jacob and Nakano, Reiichiro and others},
  journal={arXiv preprint arXiv:2110.14168},
  year={2021}
}

@article{gpqa,
  title={Gpqa: A graduate-level google-proof q\&a benchmark},
  author={Rein, David and Hou, Betty Li and Stickland, Asa Cooper and Petty, Jackson and Pang, Richard Yuanzhe and Dirani, Julien and Michael, Julian and Bowman, Samuel R},
  journal={arXiv preprint arXiv:2311.12022},
  year={2023}
}

@inproceedings{drop,
  title={DROP: A reading comprehension benchmark requiring discrete reasoning over paragraphs},
  author={Dua, Dheeru and Wang, Yizhong and Dasigi, Pradeep and Stanovsky, Gabriel and Singh, Sameer and Gardner, Matt},
  booktitle={Proceedings of the 2019 Conference of the North American Chapter of the Association for Computational Linguistics: Human Language Technologies, Volume 1 (Long and Short Papers)},
  pages={2368--2378},
  year={2019}
}

@article{coqa,
  title={Coqa: A conversational question answering challenge},
  author={Reddy, Siva and Chen, Danqi and Manning, Christopher D},
  journal={Transactions of the Association for Computational Linguistics},
  volume={7},
  pages={249--266},
  year={2019},
  publisher={MIT Press One Rogers Street, Cambridge, MA 02142-1209, USA journals-info~…}
}

@inproceedings{bbh,
  title={Challenging big-bench tasks and whether chain-of-thought can solve them},
  author={Suzgun, Mirac and Scales, Nathan and Sch{\"a}rli, Nathanael and Gehrmann, Sebastian and Tay, Yi and Chung, Hyung Won and Chowdhery, Aakanksha and Le, Quoc and Chi, Ed H and Zhou, Denny and others},
  booktitle={Findings of the Association for Computational Linguistics: ACL 2023},
  pages={13003--13051},
  year={2023}
}

@article{babi,
  title={Towards ai-complete question answering: A set of prerequisite toy tasks},
  author={Weston, Jason and Bordes, Antoine and Chopra, Sumit and Rush, Alexander M and Van Merri{\"e}nboer, Bart and Joulin, Armand and Mikolov, Tomas},
  journal={arXiv preprint arXiv:1502.05698},
  year={2015}
}

@article{arc,
  title={Think you have solved question answering? try arc, the ai2 reasoning challenge},
  author={Clark, Peter and Cowhey, Isaac and Etzioni, Oren and Khot, Tushar and Sabharwal, Ashish and Schoenick, Carissa and Tafjord, Oyvind},
  journal={arXiv preprint arXiv:1803.05457},
  year={2018}
}

@ARTICLE{logiqa2,
  author={Liu, Hanmeng and Liu, Jian and Cui, Leyang and Teng, Zhiyang and Duan, Nan and Zhou, Ming and Zhang, Yue},
  journal={IEEE/ACM Transactions on Audio, Speech, and Language Processing}, 
  title={LogiQA 2.0—An Improved Dataset for Logical Reasoning in Natural Language Understanding}, 
  year={2023},
  volume={31},
  number={},
  pages={2947-2962},
  doi={10.1109/TASLP.2023.3293046}}

@article{mbpp,
  title={Program synthesis with large language models},
  author={Austin, Jacob and Odena, Augustus and Nye, Maxwell and Bosma, Maarten and Michalewski, Henryk and Dohan, David and Jiang, Ellen and Cai, Carrie and Terry, Michael and Le, Quoc and others},
  journal={arXiv preprint arXiv:2108.07732},
  year={2021}
}

@article{math,
  title={Measuring mathematical problem solving with the math dataset},
  author={Hendrycks, Dan and Burns, Collin and Kadavath, Saurav and Arora, Akul and Basart, Steven and Tang, Eric and Song, Dawn and Steinhardt, Jacob},
  journal={arXiv preprint arXiv:2103.03874},
  year={2021}
}

@inproceedings{mmbench,
  title={Mmbench: Is your multi-modal model an all-around player?},
  author={Liu, Yuan and Duan, Haodong and Zhang, Yuanhan and Li, Bo and Zhang, Songyang and Zhao, Wangbo and Yuan, Yike and Wang, Jiaqi and He, Conghui and Liu, Ziwei and others},
  booktitle={European conference on computer vision},
  pages={216--233},
  year={2024},
  organization={Springer}
}

@inproceedings{mmmu,
  title={Mmmu: A massive multi-discipline multimodal understanding and reasoning benchmark for expert agi},
  author={Yue, Xiang and Ni, Yuansheng and Zhang, Kai and Zheng, Tianyu and Liu, Ruoqi and Zhang, Ge and Stevens, Samuel and Jiang, Dongfu and Ren, Weiming and Sun, Yuxuan and others},
  booktitle={Proceedings of the IEEE/CVF conference on computer vision and pattern recognition},
  pages={9556--9567},
  year={2024}
}

@article{mmstar,
  title={Are we on the right way for evaluating large vision-language models?},
  author={Chen, Lin and Li, Jinsong and Dong, Xiaoyi and Zhang, Pan and Zang, Yuhang and Chen, Zehui and Duan, Haodong and Wang, Jiaqi and Qiao, Yu and Lin, Dahua and others},
  journal={Advances in Neural Information Processing Systems},
  volume={37},
  pages={27056--27087},
  year={2024}
}

@inproceedings{pope,
  title={Evaluating object hallucination in large vision-language models},
  author={Li, Yifan and Du, Yifan and Zhou, Kun and Wang, Jinpeng and Zhao, Xin and Wen, Ji-Rong},
  booktitle={Proceedings of the 2023 conference on empirical methods in natural language processing},
  pages={292--305},
  year={2023}
}

@misc{realworldqa,
  title        = {Grok-1.5 Vision Preview},
  author       = {{xAI}},
  year         = {2024},
  month        = apr,
  howpublished = {\url{https://x.ai/news/grok-1.5v}},
  note         = {Introduces the RealWorldQA benchmark}
}

@misc{codealpaca,
  author = {Sahil Chaudhary},
  title = {Code Alpaca: An Instruction-following LLaMA model for code generation},
  year = {2023},
  publisher = {GitHub},
  journal = {GitHub repository},
  howpublished = {\url{https://github.com/sahil280114/codealpaca}},
}

@article{yu2023metamath,
  title={Metamath: Bootstrap your own mathematical questions for large language models},
  author={Yu, Longhui and Jiang, Weisen and Shi, Han and Yu, Jincheng and Liu, Zhengying and Zhang, Yu and Kwok, James T and Li, Zhenguo and Weller, Adrian and Liu, Weiyang},
  journal={arXiv preprint arXiv:2309.12284},
  year={2023}
}

@misc{DatabricksBlog2023DollyV2,
    author    = {Mike Conover and Matt Hayes and Ankit Mathur and Jianwei Xie and Jun Wan and Sam Shah and Ali Ghodsi and Patrick Wendell and Matei Zaharia and Reynold Xin},
    title     = {Free Dolly: Introducing the World's First Truly Open Instruction-Tuned LLM},
    year      = {2023},
    url       = {https://www.databricks.com/blog/2023/04/12/dolly-first-open-commercially-viable-instruction-tuned-llm},
    urldate   = {2023-06-30}
}

@inproceedings{li2023halueval,
  title={Halueval: A large-scale hallucination evaluation benchmark for large language models},
  author={Li, Junyi and Cheng, Xiaoxue and Zhao, Xin and Nie, Jian-Yun and Wen, Ji-Rong},
  booktitle={The 2023 Conference on Empirical Methods in Natural Language Processing},
  year={2023}
}

\newpage
\appendix

\section{Method Details}
\label{sec:appendix}
\subsection{Overall Algorithm}
The complete pseudocode of our NeuPAT method is shown in Algorithm~\ref{alg:NeuPAT}.

\begin{algorithm}[t]
\caption{NeuPAT Training Procedure}
\label{alg:NeuPAT}
\begin{algorithmic}[1]
\Require Aligned model $\Theta^0$, probing sets $\mathcal{D}_T,\mathcal{D}_V$,
training set $\mathcal{D}_{\mathrm{ori}}$, thresholds $\tau_T,\tau_V$
\Ensure Tuned model $\Theta$

\State Compute text and vision response scores
$\{s^T_{l,u},s^V_{l,u}\}$ using $\mathcal{D}_T$ and $\mathcal{D}_V$

\For{each layer $l$}
    \State Select important sets $\mathcal{T}_l$ and $\mathcal{V}_l$
    using cumulative importance thresholds $\tau_T$ and $\tau_V$
    \State $\mathcal{C}^{text}_l
    \gets \mathcal{T}_l\setminus\mathcal{V}_l$
    \State $\mathcal{C}^{vision}_l
    \gets \mathcal{V}_l\setminus\mathcal{T}_l$
    \State $\mathcal{C}^{high}_l
    \gets \mathcal{T}_l\cap\mathcal{V}_l$
    \State $\mathcal{C}^{low}_l
    \gets \overline{\mathcal{T}_l\cup\mathcal{V}_l}$
\EndFor

\State $\Theta\gets\Theta^0$
\For{each minibatch $\mathcal{B}\subset\mathcal{D}_{\mathrm{ori}}$}
    \State Compute $\mathcal{L}_{\mathrm{ori}}$ and
    $\mathcal{L}=\mathcal{L}_{\mathrm{ori}}+\mathcal{R}_{\mathrm{high}}$
    \State Freeze text neurons, update vision and low response neurons with
    $\mathcal{L}_{\mathrm{ori}}$, and update high response neurons with $\mathcal{L}$
\EndFor

\State \Return $\Theta$
\end{algorithmic}
\end{algorithm}

\begin{table}[t]
\centering
\caption{Training configurations for the two-stage pipeline.}
\label{tab:training_details}
\renewcommand{\arraystretch}{1.10} 
\resizebox{\columnwidth}{!}{
\begin{tabular}{lcc}
\toprule
\textbf{Configuration}
& \textbf{\makecell[c]{Stage 1: \\ Alignment}} 
& \textbf{\makecell[c]{Stage 2: \\ Instruction Tuning}} \\ 
\midrule
Dataset
& LLaVA-558K
& LLaVA-NeXT-780K \\

Training steps
& 2,500
& 3,500 \\

Trainable modules
& Adapter only
& \makecell[c]{LLM, adapter, \\ and vision encoder} \\ 

Global batch size
& 8
& 224 \\

Micro-batch size
& 1
& 1 \\

Gradient accumulation
& 1
& 28 \\

Number of GPUs
& 8
& 8 \\

Peak learning rate
& $1\times10^{-4}$
& $1\times10^{-5}$ \\

Minimum learning rate
& $1\times10^{-6}$
& $1\times10^{-6}$ \\

Warmup ratio
& 0.002
& 0.002 \\

Optimizer
& \multicolumn{2}{c}{\makecell[c]{Adam ($\beta_1=0.9$, $\beta_2=0.99$, \\ $\epsilon=10^{-5}$)}} \\ 

LR scheduler
& \multicolumn{2}{c}{Cosine decay} \\

Weight decay
& \multicolumn{2}{c}{0} \\

Gradient clipping
& \multicolumn{2}{c}{1.0} \\

Precision
& \multicolumn{2}{c}{BF16} \\

Sequence length
& \multicolumn{2}{c}{32,768} \\

Tensor / pipeline parallelism
& \multicolumn{2}{c}{1 / 1} \\

Image resolution
& Default
& 1,000 \\

Offline packing
& Enabled
& Disabled \\
\bottomrule
\end{tabular}%
}
\end{table}

\begin{table*}[t]
\centering
\caption{
Sensitivity analysis of the target importance mass  $\tau_a$ on language and
multimodal benchmarks. The selected threshold $\tau_a=0.8$ is highlighted.
}
\label{tab:energy_threshold}
\renewcommand{\arraystretch}{1.10}
\setlength{\tabcolsep}{2.2pt}
\scriptsize

\resizebox{\textwidth}{!}{
\begin{tabular}{
c|
ccccccccccc c|
ccccc c
}
\toprule
\multirow{2}{*}{$\boldsymbol{\tau_a}$}
& \multicolumn{12}{c|}{\textbf{Language Benchmarks}}
& \multicolumn{6}{c}{\textbf{Multimodal Benchmarks}} \\
\cmidrule(lr){2-13}
\cmidrule(lr){14-19}

& \textbf{SIQA}
& \textbf{ARC-C}
& \textbf{GPQA}
& \textbf{MATH}
& \textbf{GSM8K}
& \textbf{LogiQA2}
& \textbf{BBH}
& \textbf{bAbI}
& \textbf{MBPP}
& \textbf{DROP}
& \textbf{CoQA}
& \textbf{Avg.}
& \textbf{MMB}
& \textbf{RWQA}
& \textbf{MMMU}
& \textbf{POPE}
& \textbf{MMStar}
& \textbf{Avg.} \\
\midrule

0.6
& 49.54
& 56.57
& 35.27
& 54.40
& 84.84
& 36.77
& 53.46
& 5.87
& 65.40
& 14.57
& 69.12
& 47.80
& 73.28
& 57.65
& 43.22
& 87.59
& 46.20
& 61.59 \\

0.7
& 50.51
& 56.40
& 36.16
& 55.60
& 85.14
& 35.37
& 53.17
& 11.88
& 66.60
& 14.37
& 69.85
& 48.64
& 72.51
& 57.52
& 43.67
& 87.04
& 45.59
& 61.27 \\

\rowcolor{gray!10}
0.8
& 50.26
& 56.91
& 37.27
& 56.40
& 85.97
& 36.39
& 53.88
& 10.67
& 66.60
& 15.74
& 69.53
& 49.06
& 72.48
& 56.99
& 42.22
& 88.72
& 44.85
& 61.05 \\

0.9
& 49.69
& 55.97
& 35.71
& 55.40
& 86.43
& 35.69
& 53.51
& 13.99
& 67.80
& 15.11
& 70.63
& 49.08
& 71.39
& 54.77
& 43.11
& 86.73
& 43.47
& 59.89 \\

\bottomrule
\end{tabular}
}
\end{table*}

\begin{table*}[t]
\centering
\caption{
Sensitivity analysis of the size of probing set on language and
multimodal benchmarks. The selected size $2048$ is highlighted.
}
\label{tab:probing_size}
\renewcommand{\arraystretch}{1.10}
\setlength{\tabcolsep}{2.2pt}
\scriptsize

\resizebox{\textwidth}{!}{
\begin{tabular}{
c|
ccccccccccc c|
ccccc c
}
\toprule
\multirow{2}{*}{$\boldsymbol{N_a}$}
& \multicolumn{12}{c|}{\textbf{Language Benchmarks}}
& \multicolumn{6}{c}{\textbf{Multimodal Benchmarks}} \\
\cmidrule(lr){2-13}
\cmidrule(lr){14-19}

& \textbf{SIQA}
& \textbf{ARC-C}
& \textbf{GPQA}
& \textbf{MATH}
& \textbf{GSM8K}
& \textbf{LogiQA2}
& \textbf{BBH}
& \textbf{bAbI}
& \textbf{MBPP}
& \textbf{DROP}
& \textbf{CoQA}
& \textbf{Avg.}
& \textbf{MMB}
& \textbf{RWQA}
& \textbf{MMMU}
& \textbf{POPE}
& \textbf{MMStar}
& \textbf{Avg.} \\
\midrule

512
& 50.05
& 57.08
& 35.94
& 53.40
& 85.97
& 36.20
& 52.85
& 10.41
& 67.00
& 15.18
& 70.45
& 48.59
& 72.39
& 57.25
& 41.33
& 87.00
& 46.17
& 60.83 \\

1024
& 50.36
& 56.83
& 35.94
& 52.00
& 84.76
& 36.07
& 54.12
& 14.89
& 67.00
& 14.17
& 70.32
& 48.77
& 71.99
& 56.73
& 43.11
& 87.28
& 44.47
& 60.72 \\

\rowcolor{gray!10}
2048
& 50.26
& 56.91
& 37.27
& 56.40
& 85.97
& 36.39
& 53.88
& 10.67
& 66.60
& 15.74
& 69.53
& 49.06
& 72.48
& 56.99
& 42.22
& 88.72
& 44.85
& 61.05 \\

4096
& 50.56
& 56.40
& 35.94
& 52.22
& 85.44
& 37.15
& 53.39
& 11.16
& 67.20
& 13.89
& 70.62
& 48.54
& 70.88
& 56.99
& 42.56
& 87.13
& 45.89
& 60.69 \\

\bottomrule
\end{tabular}
}
\end{table*}

\begin{table*}[t]
\centering
\caption{
Sensitivity analysis of the high response neuron regularization coefficients
$\lambda_{\mathrm{in}}$ and $\lambda_{\mathrm{out}}$. The selected setting
$\lambda_{\mathrm{in}}=\lambda_{\mathrm{out}}=0.1$ is highlighted.
}
\label{tab:regularization_coefficient}
\renewcommand{\arraystretch}{1.10}
\setlength{\tabcolsep}{2.2pt}
\scriptsize

\resizebox{\textwidth}{!}{
\begin{tabular}{
c|
ccccccccccc c|
ccccc c
}
\toprule
\multirow{2}{*}{$\boldsymbol{\lambda_{\mathrm{in}} =\lambda_{\mathrm{out}}}$}
& \multicolumn{12}{c|}{\textbf{Language Benchmarks}}
& \multicolumn{6}{c}{\textbf{Multimodal Benchmarks}} \\
\cmidrule(lr){2-13}
\cmidrule(lr){14-19}

& \textbf{SIQA}
& \textbf{ARC-C}
& \textbf{GPQA}
& \textbf{MATH}
& \textbf{GSM8K}
& \textbf{LogiQA2}
& \textbf{BBH}
& \textbf{bAbI}
& \textbf{MBPP}
& \textbf{DROP}
& \textbf{CoQA}
& \textbf{Avg.}
& \textbf{MMB}
& \textbf{RWQA}
& \textbf{MMMU}
& \textbf{POPE}
& \textbf{MMStar}
& \textbf{Avg.} \\
\midrule

0.05
& 50.26
& 57.42
& 36.61
& 52.60
& 85.97
& 34.41
& 52.02
& 10.09
& 67.00
& 15.58
& 69.70
& 48.33
& 72.68
& 57.91
& 42.11
& 87.19
& 46.29
& 61.24 \\

\rowcolor{gray!10}
0.1
& 50.26
& 56.91
& 37.27
& 56.40
& 85.97
& 36.39
& 53.88
& 10.67
& 66.60
& 15.74
& 69.53
& 49.06
& 72.48
& 56.99
& 42.22
& 88.72
& 44.85
& 61.05 \\

0.5
& 50.31
& 56.57
& 37.28
& 54.20
& 84.08
& 36.26
& 52.79
& 13.86
& 68.00
& 14.75
& 70.78
& 48.99
& 71.65
& 56.21
& 43.00
& 87.41
& 45.82
& 60.82 \\

1.0
& 50.41
& 56.14
& 37.05
& 55.60
& 85.90
& 36.26
& 52.30
& 13.38
& 66.60
& 13.71
& 70.85
& 48.93
& 71.65
& 55.95
& 43.22
& 87.41
& 44.87
& 60.62 \\

\bottomrule
\end{tabular}
}
\end{table*}

\section{Additional Related Work}
\subsection{Neuron-Level Analysis}
Recent studies have investigated the functional specialization of individual neurons in language and multimodal models. Fu et al.~\cite{fu2026correct} reveal that knowledge in MLLMs can be distributed across decoupled modality-specific neuron pathways and localize these neurons to improve knowledge editing under different modality inputs. Zhao et al.~\cite{zhao2026discovering} identify emotion-sensitive neurons in large audio-language models and causally validate their roles through neuron suppression and activation steering. For multimodal unlearning, Liu et al.~\cite{liu2025modality} locate neurons according to their relative importance across modalities and selectively prune them to remove targeted knowledge while preserving general model utility. MNAFT~\cite{li2026mnaft} identifies modality-relevant neurons using activation and gradient information and selectively fine-tunes them for image translation to reduce interference across languages and modalities. Qin et al.~\cite{qin2025achilles} further show that LLMs contain ultra-sparse critical neurons whose perturbation can severely impair overall language ability.

These studies associate neuron-level structures with modality-specific knowledge, task behaviors, and fundamental model capabilities, but primarily focus on knowledge editing, emotion control, unlearning, task-specific image translation, or vulnerability analysis. In contrast, we study neuron-wise adaptation dynamics during general multimodal learning. NeuPAT uses lightweight text and vision probing to estimate heterogeneous neuron responses, followed by neuron adaptation role allocation and role-aware multimodal tuning. By protecting language-sensitive neurons, promoting adaptation through vision-responsive and underutilized neurons, and regularizing neurons important to both text and vision, NeuPAT preserves pretrained language capabilities while supporting multimodal learning.
 
\section{Detailed Experimental Setup}
\subsection{Probing Sets}
We construct separate text-only and vision probing sets to estimate modality-specific neuron responses. The vision probing set contains $N_V=2048$ samples randomly drawn from LLaVA-NeXT-780K~\cite{liu2024llavanext}. The text probing set contains $N_T=2048$ prompts sampled from publicly available general-domain text datasets that are disjoint from all evaluation benchmarks. To improve coverage, we include prompts from instruction-following, factuality, and general reasoning domains, including 
CodeAlpaca-20k~\cite{codealpaca}, MetaMathQA~\cite{yu2023metamath}, databricks-dolly-15k~\cite{DatabricksBlog2023DollyV2} and HaluEval~\cite{li2023halueval}. We allocate samples evenly across data sources. When a source contains fewer samples than its assigned quota, the remaining quota is redistributed among the other sources. 

Both probing sets are used solely for forward-pass activation statistics.
They do not contribute to the training objective or parameter updates. We use
the same processor and chat-template pipeline for both sets, while providing
images only for the vision probing samples.

\subsection{Training Details}
As shown in Table~\ref{tab:training_details}, we follow a two-stage training pipeline. In Stage~1, we perform image-text alignment on LLaVA-558K for 2,500 steps, where only the multimodal adapter is optimized. We use a global batch size of 8 on 8 A100 GPUs, with a micro-batch size of 1. The peak learning rate is $1\times10^{-4}$ and is decayed to $1\times10^{-6}$ using a cosine schedule. The warmup ratio is set to $0.002$, corresponding to approximately 5 warmup steps.

In Stage~2, we conduct visual instruction tuning on LLaVA-NeXT-780K for 3,500 steps. The language model, multimodal adapter, and vision encoder are included in optimization, while NeuPAT applies plasticity-guided multimodal tuning strategies to the neurons in the language backbone. Training is performed on 8 A100 GPUs with a global batch size of 224, a micro-batch size of 1, and 28 gradient-accumulation steps. The peak learning rate is $1\times10^{-5}$, with the same minimum learning rate of $1\times10^{-6}$ and a warmup ratio of $0.002$. Both stages use Adam with $\beta_1=0.9$, $\beta_2=0.99$, and $\epsilon=10^{-5}$, together with cosine learning-rate decay, zero weight decay, and gradient clipping at 1.0. 

\subsection{Baseline Implementations}
\paragraph{LoRA.}
LoRA~\cite{hu2022lora} inserts trainable low-rank adapters into selected linear layers while freezing the original parameters. We set the rank to $r=32$, the scaling factor to $\alpha=64$, and the learning rate to $1\times10^{-4}$. The model is trained on the same multimodal instruction data as Vanilla Tuning.

\paragraph{EWC.}
EWC~\cite{kirkpatrick2017overcoming} estimates parameter importance using the diagonal Fisher information and penalizes changes to important parameters. We estimate the Fisher information at the shared Stage-1 checkpoint using 2,048 text-only instruction samples that are disjoint from all evaluation benchmarks. The Fisher statistics are averaged over samples, and the EWC penalty is applied only to the language-model parameters. We search $\lambda_{\mathrm{EWC}}$ over $\{0.1,1,10,100\}$ and select $\lambda_{\mathrm{EWC}}=\text{10}$ according to the text-multimodal
performance. All remaining training settings are identical to Vanilla Tuning.

\paragraph{WINGS.}
WINGS~\cite{zhang2024wings} introduces parallel visual and textual learners into attention layers to reduce over-reliance on visual tokens. Their outputs are fused with the original attention output through a learned router. We reproduce WINGS using the architecture and hyperparameter settings recommended in the original paper.

\paragraph{TIES.}
TIES~\cite{ratzlaff2024training} is a training-free merging method that sparsifies the task vector between the visually tuned model and the original LLM. We follow the original paper for the task-vector density, merge coefficient, and merging procedure.

\paragraph{Locate-then-Merge.}
Locate-then-Merge~\cite{yu2025locate} identifies high-impact neurons from parameter changes, suppresses low-impact updates, and restores selected neurons through replacement or rescaling. We follow the original paper for the neuron-retention ratio, parameter-sparsification ratio, restoration strategy, and associated coefficients. The hyperparameters are selected from the recommended ranges reported by the authors.

\paragraph{PlaM.}
PlaM~\cite{wang2026plam} locates a plateau layer through layer-wise vision-token masking and merges subsequent layers with the original LLM. We follow the original paper to determine the plateau layer and search the merge coefficients within its recommended range. Earlier visual-alignment layers remain unchanged, while the selected later layers are linearly merged with the original language backbone.

\begin{table*}[t]
\centering
\caption{
Complete cross-backbone results on 11 language and 5 multimodal benchmarks.
``LLM'' denotes the original language backbone before multimodal training.
}
\label{tab:complete_backbone_results}
\renewcommand{\arraystretch}{1.08}
\setlength{\tabcolsep}{2.2pt}
\scriptsize

\resizebox{\textwidth}{!}{
\begin{tabular}{l|ccc|ccc|ccc|ccc|ccc}
\toprule
\multirow{2}{*}{\textbf{Task}}
& \multicolumn{3}{c|}{\textbf{Qwen3-0.6B}}
& \multicolumn{3}{c|}{\textbf{Phi-4-Mini-Instruct}}
& \multicolumn{3}{c|}{\textbf{Qwen2.5-7B-Instruct}}
& \multicolumn{3}{c|}{\textbf{Llama3.1-8B-Instruct}}
& \multicolumn{3}{c}{\textbf{Qwen2.5-14B-Instruct}} \\
\cmidrule(lr){2-4}
\cmidrule(lr){5-7}
\cmidrule(lr){8-10}
\cmidrule(lr){11-13}
\cmidrule(lr){14-16}

& \textbf{LLM}
& \textbf{Vanilla Tuning}
& \textbf{NeuPAT}
& \textbf{LLM}
& \textbf{Vanilla Tuning}
& \textbf{NeuPAT}
& \textbf{LLM}
& \textbf{Vanilla Tuning}
& \textbf{NeuPAT}
& \textbf{LLM}
& \textbf{Vanilla Tuning}
& \textbf{NeuPAT}
& \textbf{LLM}
& \textbf{Vanilla Tuning}
& \textbf{NeuPAT} \\
\midrule

\multicolumn{16}{c}{\textbf{Language Benchmarks}} \\
\midrule

MATH-500
& 14.00 & 13.80 & 13.40
& 39.20 & 32.00 & 34.00
& 44.40 & 28.00 & 35.20
& 36.80 & 28.80 & 35.80
& 43.40 & 25.60 & 39.00 \\

BBH
& 33.39 & 28.69 & 33.44
& 52.82 & 40.04 & 51.29
& 45.81 & 41.98 & 47.53
& 44.59 & 34.91 & 41.68
& 52.54 & 39.87 & 52.59 \\

bAbI
& 2.50 & 1.93 & 2.21
& 2.65 & 1.06 & 2.78
& 2.71 & 0.60 & 2.56
& 0.00 & 0.14 & 0.81
& 1.17 & 0.03 & 1.48 \\

MBPP
& 27.40 & 22.60 & 28.20
& 55.40 & 47.60 & 53.80
& 47.60 & 40.20 & 42.80
& 58.40 & 54.20 & 56.80
& 66.80 & 65.60 & 66.80 \\

LogiQA2
& 30.34 & 28.75 & 30.79
& 36.07 & 30.79 & 35.08
& 40.20 & 35.18 & 40.52
& 38.36 & 31.87 & 37.45
& 42.88 & 37.47 & 40.14 \\

GSM8K
& 40.56 & 35.18 & 40.11
& 83.62 & 73.92 & 81.45
& 76.50 & 74.83 & 78.70
& 78.17 & 67.55 & 74.55
& 79.83 & 79.30 & 85.44 \\

DROP
& 12.32 & 6.12 & 9.80
& 16.01 & 13.39 & 16.94
& 16.22 & 7.13 & 11.69
& 12.05 & 7.15 & 11.53
& 20.60 & 20.60 & 26.41 \\

CoQA
& 57.57 & 55.93 & 59.42
& 78.40 & 65.23 & 80.04
& 78.74 & 73.63 & 78.80
& 77.83 & 77.50 & 80.48
& 78.20 & 74.44 & 77.51 \\

SocialIQA
& 40.38 & 39.00 & 40.33
& 49.59 & 46.11 & 50.06
& 51.59 & 50.20 & 55.89
& 49.85 & 49.74 & 49.95
& 54.04 & 51.23 & 55.22 \\

GPQA
& 29.69 & 27.90 & 29.12
& 30.58 & 30.13 & 31.70
& 34.38 & 32.37 & 34.38
& 34.60 & 31.03 & 33.81
& 36.83 & 35.04 & 36.16 \\

ARC-C
& 34.39 & 34.04 & 36.09
& 58.45 & 58.36 & 58.70
& 55.29 & 52.65 & 54.10
& 53.41 & 53.24 & 53.33
& 60.41 & 60.41 & 61.26 \\

\midrule
\textbf{Text Avg.}
& \textbf{29.32} & \textbf{26.72} & \textbf{29.36}
& \textbf{45.71} & \textbf{39.88} & \textbf{45.08}
& \textbf{44.86} & \textbf{39.71} & \textbf{43.83}
& \textbf{44.01} & \textbf{39.65} & \textbf{43.29}
& \textbf{48.79} & \textbf{44.51} & \textbf{49.27} \\

\midrule
\multicolumn{16}{c}{\textbf{Multimodal Benchmarks}} \\
\midrule

MMBench-EN
& -- & 52.58 & 52.52
& -- & 50.52 & 50.89
& -- & 64.78 & 64.64
& -- & 60.14 & 61.94
& -- & 75.17 & 74.31 \\

RealWorldQA
& -- & 45.10 & 44.75
& -- & 45.10 & 44.31
& -- & 52.03 & 52.01
& -- & 26.67 & 44.97
& -- & 56.34 & 55.64 \\

MMMU
& -- & 30.67 & 32.67
& -- & 36.78 & 36.22
& -- & 41.00 & 42.56
& -- & 35.89 & 36.22
& -- & 47.67 & 48.11 \\

POPE
& -- & 85.42 & 84.26
& -- & 77.16 & 77.82
& -- & 87.12 & 86.54
& -- & 82.23 & 80.90
& -- & 87.27 & 87.67 \\

MMStar
& -- & 37.23 & 37.07
& -- & 30.58 & 31.26
& -- & 40.56 & 40.37
& -- & 36.54 & 33.94
& -- & 50.26 & 50.14 \\

\midrule
\textbf{MM Avg.}
& -- & \textbf{50.20} & \textbf{50.25}
& -- & \textbf{48.03} & \textbf{48.10}
& -- & \textbf{57.10} & \textbf{57.22}
& -- & \textbf{48.29} & \textbf{51.59}
& -- & \textbf{63.34} & \textbf{63.17} \\

\bottomrule
\end{tabular}
}
\end{table*}

\section{Additional Experimental Results}
\subsection{Sensitivity Analysis}

All sensitivity analyses are conducted after
fixing the default configuration used in the main experiments. We vary one
hyperparameter at a time while keeping all other settings unchanged. These
experiments are intended to evaluate robustness.

\paragraph{Target Importance Mass $\tau_a$.}
We analyze the sensitivity to the target importance mass $\tau_a$ used in
neuron adaptation role allocation. As shown in
Table~\ref{tab:energy_threshold}, language performance generally improves as
$\tau_a$ increases. A larger $\tau_a$ retains more cumulative response mass
for each modality, thereby expanding the selected neuron sets and reducing
the proportion of low response neurons. This tends to assign more neurons to
text-related or high response neurons, strengthening language preservation but
leaving less unconstrained capacity for multimodal adaptation. Consequently,
multimodal performance gradually declines at larger values of $\tau_a$. The
results support our default choice of $\tau_a=0.8$, which achieves a favorable
balance between language preservation and multimodal adaptation.

\paragraph{Probing Set Size.}
We further examine the sensitivity of NeuPAT to the number of samples in each
probing set. As shown in Table~\ref{tab:probing_size}, the overall performance
is relatively stable across different sizes, with variations of only $0.52$
and $0.36$ points in the text and multimodal averages, respectively.
Increasing $N_a$ from $512$ to $2048$ generally improves performance, and
$N_a=2048$ achieves the best averages on both language and multimodal
benchmarks. Further increasing the size to $4096$ provides no additional
benefit, indicating diminishing returns from additional probing samples. The
results support $N_a=2048$ as a reasonable default while showing that NeuPAT
is relatively insensitive to the probing-set size.

\paragraph{High Response Neuron Regularization Coefficients.}
We study the sensitivity to the high response neuron regularization coefficients by
setting $\lambda_{\mathrm{in}}=\lambda_{\mathrm{out}}$. As shown in
Table~\ref{tab:regularization_coefficient}, a weak constraint of $0.05$
achieves the highest multimodal average but lower language performance.
Increasing the coefficients beyond $0.1$ provides no further text improvement
and gradually reduces multimodal performance. These results support our
default setting of
$\lambda_{\mathrm{in}}=\lambda_{\mathrm{out}}=0.1$, which provides a favorable
text-multimodal trade-off.

\begin{table*}[t]
\centering
\caption{
Complete ablation results on 11 language and 5 multimodal benchmarks.
Variants are grouped by global update strategy, neuron allocation, and
neuron-wise plasticity allocation strategies. The complete NeuPAT configuration is repeated
and highlighted in each group for comparison.
}
\label{tab:complete_ablation_results}
\renewcommand{\arraystretch}{1.10}
\setlength{\tabcolsep}{2.2pt}
\scriptsize

\resizebox{\textwidth}{!}{
\begin{tabular}{
l|
ccccccccccc c|
ccccc c
}
\toprule
\multirow{2}{*}{\textbf{Variant}}
& \multicolumn{12}{c|}{\textbf{Language Benchmarks}}
& \multicolumn{6}{c}{\textbf{Multimodal Benchmarks}} \\
\cmidrule(lr){2-13}
\cmidrule(lr){14-19}

& \textbf{SIQA}
& \textbf{ARC-C}
& \textbf{GPQA}
& \textbf{MATH}
& \textbf{GSM8K}
& \textbf{LogiQA2}
& \textbf{BBH}
& \textbf{bAbI}
& \textbf{MBPP}
& \textbf{DROP}
& \textbf{CoQA}
& \textbf{Avg.}
& \textbf{MMB}
& \textbf{RWQA}
& \textbf{MMMU}
& \textbf{POPE}
& \textbf{MMStar}
& \textbf{Avg.} \\
\midrule

\multicolumn{19}{l}{
\textit{(a) Global vs. Neuron-Aware Update}
} \\
\addlinespace[1pt]

Global Freeze
& 49.33 & 56.06 & 37.72 & 61.20 & 86.05 & 34.29
& 53.57 & 13.65 & 67.10 & 14.90 & 70.37 & 49.48
& 68.64 & 53.99 & 41.33 & 86.03 & 43.08 & 58.61 \\

Uniform Update (Vanilla Tuning)
& 46.78 & 55.72 & 34.15 & 50.40 & 83.47 & 32.95
& 37.12 & 2.68 & 62.60 & 9.90 & 67.37 & 43.92
& 72.08 & 56.60 & 43.11 & 87.69 & 45.40 & 60.98 \\

Global Reg.
& 49.33 & 56.14 & 36.83 & 54.40 & 84.99 & 34.73
& 55.23 & 9.33 & 65.50 & 9.68 & 70.15 & 47.85
& 72.08 & 54.25 & 42.56 & 86.03 & 43.91 & 59.77 \\

\rowcolor{gray!10}
\textbf{NeuPAT}
& 50.26 & 56.91 & 37.27 & 56.40 & 85.97 & 36.39
& 53.88 & 10.67 & 66.60 & 15.74 & 69.53 & 49.06
& 72.48 & 56.99 & 42.22 & 88.72 & 44.85 & 61.05 \\

\midrule

\multicolumn{19}{l}{
\textit{(b) Neuron Allocation Strategy}
} \\
\addlinespace[1pt]

Random Partition
& 49.33 & 55.55 & 36.16 & 55.40 & 83.69 & 34.41
& 54.32 & 1.94 & 65.40 & 9.67 & 69.42 & 46.84
& 69.76 & 54.51 & 43.33 & 87.52 & 44.47 & 59.92 \\

Fixed-ratio Partition
& 50.26 & 56.23 & 35.49 & 53.20 & 84.90 & 35.94
& 52.85 & 10.24 & 66.20 & 10.11 & 69.85 & 47.75
& 72.16 & 57.25 & 41.33 & 87.28 & 43.26 & 60.26 \\

\rowcolor{gray!10}
\textbf{NeuPAT}
& 50.26 & 56.91 & 37.27 & 56.40 & 85.97 & 36.39
& 53.88 & 10.67 & 66.60 & 15.74 & 69.53 & 49.06
& 72.48 & 56.99 & 42.22 & 88.72 & 44.85 & 61.05 \\

\midrule

\multicolumn{19}{l}{
\textit{(c) Neuron-Wise Update Constraint}
} \\
\addlinespace[1pt]

w/o Text Freeze
& 49.80 & 55.29 & 35.71 & 52.60 & 83.90 & 36.64
& 52.96 & 5.87 & 64.80 & 10.39 & 70.23 & 47.11
& 72.11 & 57.30 & 42.33 & 87.20 & 45.68 & 60.92 \\

w/o Vision Update
& 50.41 & 55.63 & 36.16 & 53.20 & 84.44 & 36.45
& 53.99 & 11.85 & 67.10 & 12.77 & 70.20 & 48.38
& 69.69 & 55.16 & 42.44 & 86.61 & 44.04 & 59.59 \\

w/o Low Response Update
& 49.74 & 56.31 & 37.28 & 53.00 & 84.76 & 35.88
& 53.89 & 11.64 & 66.80 & 9.96 & 70.13 & 48.13
& 71.05 & 56.34 & 42.67 & 86.94 & 45.72 & 60.54 \\

High Response Full Update
& 48.41 & 55.46 & 34.15 & 53.60 & 83.61 & 37.47
& 52.40 & 9.94 & 64.00 & 11.99 & 67.82 & 47.17
& 72.34 & 57.25 & 42.78 & 87.27 & 44.72 & 60.87 \\

High Response Freeze
& 49.90 & 56.66 & 37.05 & 55.80 & 85.06 & 35.37
& 53.29 & 15.16 & 67.20 & 13.70 & 70.48 & 49.06
& 69.93 & 55.03 & 42.67 & 87.30 & 45.10 & 60.01 \\

\rowcolor{gray!10}
\textbf{NeuPAT}
& 50.26 & 56.91 & 37.27 & 56.40 & 85.97 & 36.39
& 53.88 & 10.67 & 66.60 & 15.74 & 69.53 & 49.06
& 72.48 & 56.99 & 42.22 & 88.72 & 44.85 & 61.05 \\

\midrule

\multicolumn{19}{l}{
\textit{(d) High Response Neuron Regularization Design}
} \\
\addlinespace[1pt]

l2-l2
& 50.00 & 55.57 & 37.95 & 55.20 & 85.22 & 36.13
& 52.97 & 11.41 & 66.20 & 13.06 & 69.45 & 48.47
& 72.20 & 57.65 & 42.00 & 87.84 & 46.07 & 61.15 \\

cos-cos
& 50.05 & 56.66 & 37.28 & 54.20 & 84.99 & 36.51
& 52.34 & 8.44 & 67.40 & 14.51 & 69.62 & 48.36
& 72.16 & 56.99 & 42.78 & 87.37 & 44.50 & 60.76 \\

cos-l2
& 50.00 & 56.91 & 36.38 & 54.60 & 85.22 & 36.07
& 53.26 & 15.60 & 67.00 & 13.40 & 69.75 & 48.93
& 72.51 & 55.29 & 43.44 & 87.14 & 45.26 & 60.73 \\

\rowcolor{gray!10}
\textbf{NeuPAT (l2-cos)}
& 50.26 & 56.91 & 37.27 & 56.40 & 85.97 & 36.39
& 53.88 & 10.67 & 66.60 & 15.74 & 69.53 & 49.06
& 72.48 & 56.99 & 42.22 & 88.72 & 44.85 & 61.05 \\

\bottomrule
\end{tabular}
}
\end{table*}

\begin{figure*}[t]
\centering
\includegraphics[width=0.99\textwidth]{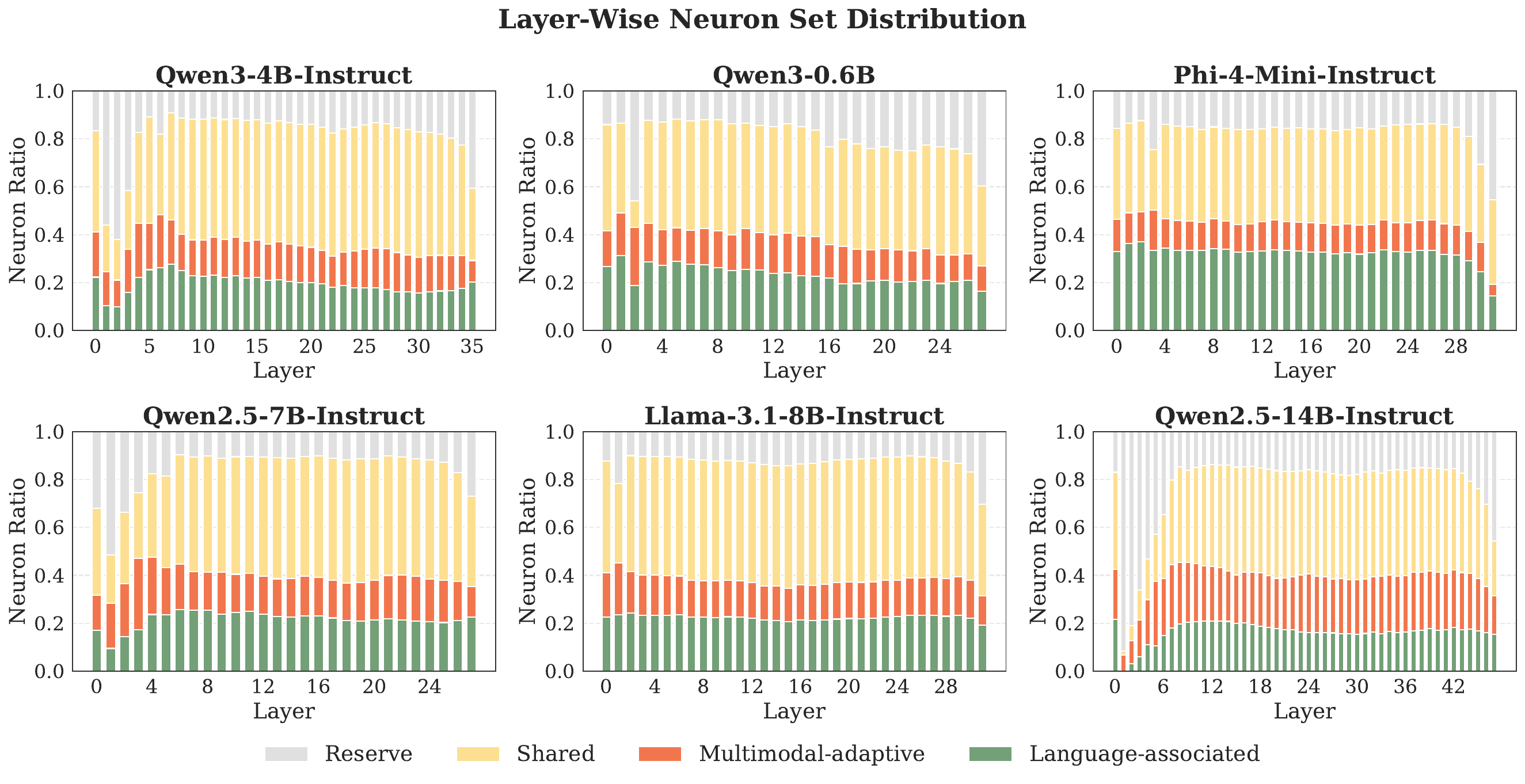} 
\caption{\textbf{Layer-wise neuron distributions across LLM backbones.}
High response neurons form the largest group in most layers, while the others vary across models and depths.}
\label{fig:ffn_neuron_category}
\end{figure*}

\subsection{Complete Cross-Backbone Results}

Due to space limitations, the main text reports results on five representative language benchmarks. Table~\ref{tab:complete_backbone_results} provides the complete results on all 11 language and 5 multimodal benchmarks. Vanilla Tuning consistently degrades the average language performance across all tested language backbones. In contrast, NeuPAT improves over Vanilla Tuning by $2.64$, $5.20$, $4.12$, $3.64$, and $4.76$ points on Qwen3-0.6B, Phi-4-Mini-Instruct, Qwen2.5-7B-Instruct, Llama3.1-8B-Instruct, and Qwen2.5-14B-Instruct, respectively. It also recovers the original LLM average within approximately one point for all backbones and slightly surpasses it on Qwen3-0.6B and Qwen2.5-14B-Instruct. Meanwhile, multimodal performance remains comparable to Vanilla Tuning, with particularly notable gains on Llama3.1-8B-Instruct. Although the improvements vary across individual tasks, the overall results confirm that NeuPAT generalizes across different model families and scales.

\subsection{Complete Ablation Results}
\paragraph{Complete Results for the Main-Text Ablations.}
Table~\ref{tab:complete_ablation_results} reports the complete results for the three ablation groups presented in the main text. Global update strategies reveal the trade-off between language preservation and multimodal adaptation. Alternative partitioning strategies verify the importance of neuron adaptation role allocation, while the update-strategy ablations demonstrate the role of each neuron set. The complete results are consistent with the conclusions drawn from the representative benchmarks in the main text.

\paragraph{High Response Neuron Regularization Design.}
We further compare different input- and output-side regularization combinations for high response neurons. As shown in Table~\ref{tab:complete_ablation_results}(d), l2-l2 achieves slightly higher multimodal performance but lower text performance, while cos-cos and cos-l2 yield weaker overall trade-offs. The proposed l2-cos design obtains the highest text average while maintaining near-best multimodal performance, providing the most balanced result. This suggests that input-side parameters benefit from magnitude constraints, whereas output-side parameters are better regularized by preserving their transformation directions.

\subsection{Additional Neuron Visualizations}

\subsubsection{Neuron Distribution across LLM Backbones}
Figure~\ref{fig:ffn_neuron_category} compares the layer-wise distributions of the neuron sets identified by importance-guided allocation across different LLM backbones. Although their proportions vary across model families, scales, and layers, text-critical, vision-critical, high response, and low response neurons consistently coexist throughout the backbone, with high response neurons generally forming the largest set. Some models exhibit stronger fluctuations in early layers, while later layers tend to show more stable distributions. These consistent yet heterogeneous patterns suggest that neuron-wise adaptation dynamics generalize across backbones, supporting the architecture-agnostic applicability of NeuPAT.

\end{document}